\documentclass[10pt,journal,compsoc]{IEEEtran}
\ifCLASSOPTIONcompsoc
  \usepackage[nocompress]{cite}
\else
  \usepackage{cite}
\fi

\ifCLASSINFOpdf
\else
\fi

\usepackage{graphicx}
\usepackage{amsmath}
\usepackage{amsfonts}
\usepackage{bm}
\usepackage{makecell}
\usepackage{indentfirst,booktabs,multirow,tabularx}
\usepackage[table]{xcolor}
\usepackage[noend]{algpseudocode}
\usepackage{algorithmicx,algorithm}
\usepackage[colorlinks,bookmarksopen,bookmarksnumbered,citecolor=green, linkcolor=blue, urlcolor=red]{hyperref}
\usepackage{subfigure}

\usepackage{balance}

\definecolor{OverviewDatabase}{RGB}{0,176,80}
\definecolor{OverviewMethod}{RGB}{0,112,192}
\definecolor{Checked}{RGB}{0,176,240}

\definecolor{TableLine}{RGB}{234,234,254}
\definecolor{ltgray}{rgb}{0.84,0.84,0.84}
\definecolor{ltblue}{rgb}{0.356,0.608,0.835}

\begin{document}
%
\title{Blind Multimodal Quality Assessment of Low-light Images}
\author{Miaohui~Wang,~\IEEEmembership{Member,~IEEE,}
		Zhuowei~Xu,
		Mai~Xu,~\IEEEmembership{Senior Member,~IEEE,}
        and Weisi~Lin,~\IEEEmembership{Fellow,~IEEE,}

\IEEEcompsocitemizethanks{\IEEEcompsocthanksitem M. Wang and Z. Xu are with the Guangdong Key Laboratory of Intelligent Information Processing, Shenzhen University, Shenzhen, China.\protect\\E-mail: wang.miaohui@gmail.com
\IEEEcompsocthanksitem  M. Xu is with the School of Electronic and Information Engineering, Beihang University, Beijing. E-mail: maixu@buaa.edu.cn.
\IEEEcompsocthanksitem  W. Lin is with the School of Computer Science and Engineering, Nanyang Technological University, 50 Nanyang Avenue, Singapore. \protect\\E-mail: wslin@ntu.edu.sg.
}
}

\markboth{~~~}%
{Wang \MakeLowercase{\textit{et al.}}: Blind Multimodal Quality Assessment of Low-light Images}
%


\IEEEtitleabstractindextext{%
\begin{abstract}
Blind image quality assessment (BIQA) aims at automatically and accurately forecasting objective scores for visual signals, which has been widely used to monitor product and service quality in low-light applications, covering smartphone photography, video surveillance, autonomous driving, \textit{etc}. Recent developments in this field are dominated by unimodal solutions inconsistent with human subjective rating patterns, where human visual perception is simultaneously reflected by multiple sensory information. 
In this article, we present a unique blind multimodal quality assessment (BMQA) of low-light images from subjective evaluation to objective score. 
To investigate the multimodal mechanism, we first establish a multimodal low-light image quality (MLIQ) database with authentic low-light distortions, containing image-text modality pairs. Further, we specially design the key modules of BMQA, considering multimodal quality representation, latent feature alignment and fusion, and hybrid self-supervised and supervised learning. 
Extensive experiments show that our BMQA yields state-of-the-art accuracy on the proposed MLIQ benchmark database. In particular, we also build an independent single-image modality Dark-4K database, which is used to verify its applicability and generalization performance in mainstream unimodal applications. Qualitative and quantitative results on Dark-4K show that BMQA achieves superior performance to existing BIQA approaches as long as a pre-trained model is provided to generate text description. 
The proposed framework and two databases as well as the collected BIQA methods and evaluation metrics are made publicly available on here. 
\end{abstract}

}

\maketitle

\IEEEdisplaynontitleabstractindextext

%
\IEEEpeerreviewmaketitle

\IEEEraisesectionheading{\section{Introduction}\label{sec:introduction}}

\IEEEPARstart{S}{torage}, transmission, and processing of low-light images are unavoidable \cite{li2021low}, especially in smartphone photography, video surveillance, autonomous driving, \textit{etc}. However, imaging in weak-illumination environments can lead to uneven brightness, poor visibility, impaired color, and increased hybrid noise, which degrade user experience and product value \cite{wang2022low}. Further, low-light images also pose various challenges to the performance of mainstream vision algorithms, including object detection \cite{cui2021multitask}, recognition \cite{xu2022aligning}, classification \cite{loh2019getting}, tracking \cite{ye2021darklighter}, assessment \cite{wang2022super},  segmentation  \cite{deng2022nightlab}, and enhancement \cite{zhang2021beyond}. Therefore, it is essential to develop a reliable objective quality indicator for low-light images, which helps to meet the quality measurement and inspection needs in various industrial products and computer vision tasks.

Low-light blind image quality assessment (BIQA) aims to automatically and accurately estimate objective scores, thereby avoiding the obstacles of subjective experiments such as time-consuming, unstable, and non-automated processes. This is particularly important for quality monitoring in industrial products \cite{wang2022low}.  At the same time, the human visual system (HVS) is the ultimate receiver of visual signals in the BIQA task \cite{zhang2022continual}, and human visual perception is simultaneously reflected by multiple sensory information. However, existing BIQA methods, whether hand-crafted or deep-learned, rarely consider multimodal information \cite{min2020study} and are limited to low-light images alone. As a result, how to utilize multimodal learning to more accurately perform the quality assessment of low-light images is the most fundamental motivation behind this work.

The perception of image quality is a complex and subjective process that involves evaluating and interpreting visual stimuli. When scoring the quality of visual signals, humans can perceive multiple sensing information at the same time \cite{pinson2012influence,baltruvsaitis2018multimodal}. 
After acquired exercise, our brains can easily make connections between different modality data and further create a comprehensive representation of the characteristics of things \cite{song2019harmonized}. For example, when the image modality is influenced by various low-light noises, other auxiliary modalities are expected to provide supplementary quality description clues, such as the text description of image content or semantic visual understanding \cite{wang2021semantic}. Consequently, multimodal BIQA aims to create a visual indicator that mimics the HVS and learns better quality descriptors that represent human visual perception.

Inspired by the above discussion, we propose an early multimodal BIQA paradigm for low-light images. Considering that there is no low-light BIQA database equipped with multimodal information, we have constructed the first \underline{M}ultimodal \underline{L}ow-light \underline{I}mage \underline{Q}uality (MLIQ) database. In the image modality, low-light images contain authentic distortions from the steps of image acquisition and processing \cite{wang2022low}. In the text modality, we have specified quality-aware principles for generating semantic  descriptions of image quality, which are based on the fact that humans are better at describing quality cognition rather than giving a quantitative value \cite{yang2022fine}. Thus, text-based quality semantic description (QSD) can provide supplementary information in the modeling of BIQA.

Further, we have developed a unique \underline{B}lind \underline{M}ultimodal \underline{Q}uality \underline{A}ssessment (BMQA) method to integrate image and text features.  The integration of cross-modal information helps maintain the representation depth of objective visual signals while broadening the breadth of human visual perception, which can introduce new benefits for the learning of image quality descriptors. The expansion of data modality helps a deep-learned model to enrich low-level embedding features from different perspectives, thereby improving the robustness of the forecasting performance \cite{baltruvsaitis2018multimodal}. Extensive experiments validate the effectiveness of the proposed BMQA, demonstrating the great potential of multimodal learning in blind quality assessment modeling.

\begin{figure}[!t]
\centering
\includegraphics[width=1.0\linewidth, height=0.40\linewidth]{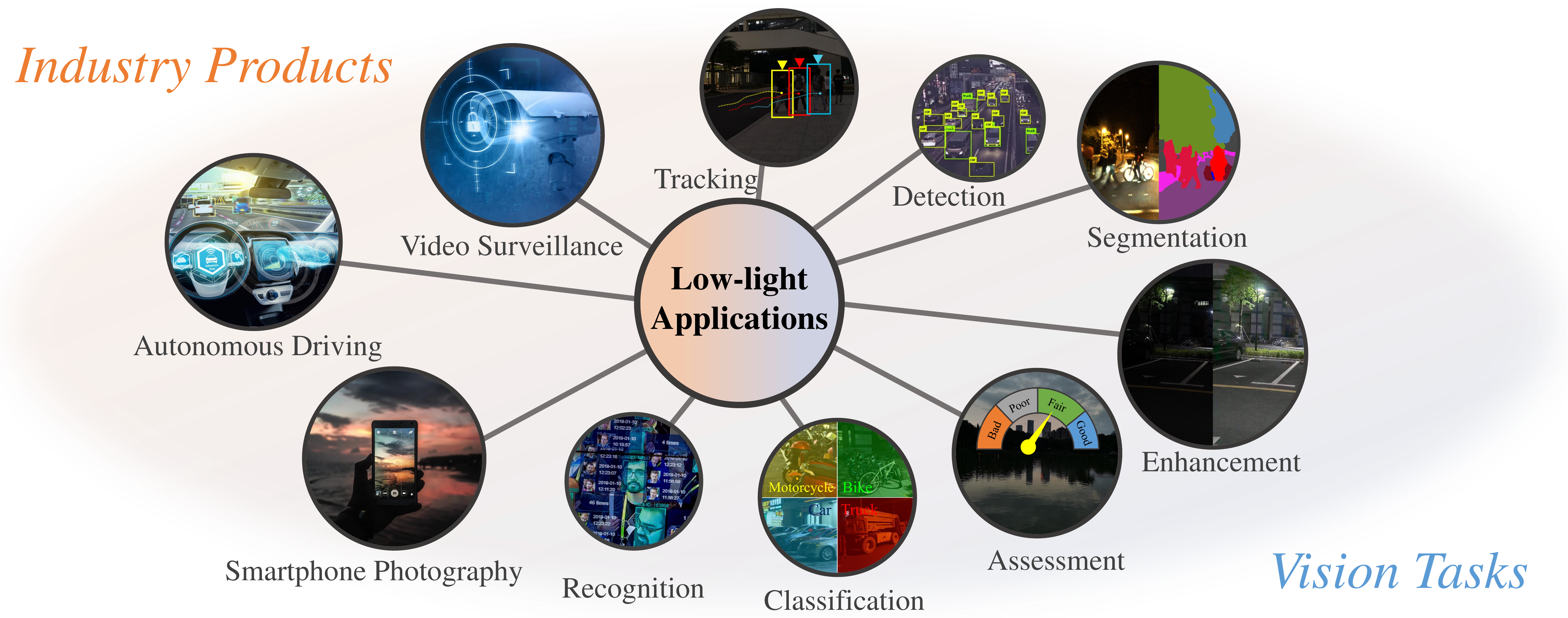}  
\caption{\textbf{Typical applications of low-light image}. Please zoom in the electronic version for better details.}
\label{fig:illustration_mliq_database}
\end{figure}

The main contributions are four-folds:
\begin{itemize}

\item Inspired by the HVS, we propose to apply multimodal learning to the BIQA problem by integrating visual and text features. To the best of our survey, this is one of the first attempts to explicitly explore low-light quality assessment across different modalities.

\item To verify the feasibility of multimodality in the BIQA task, we first  construct a new MLIQ database of low-light images, which contains $3600$ image-text data pairs. In addition, we carry out a statistical analysis of the text feature, which is helpful to demonstrate human quality cognition.

\item Based on the MLIQ database, we further investigate three main modules in multimodal learning: image-text quality representations, latent feature alignment, and fusion prediction. To improve the efficiency of deep model training, we develop an effective BMQA method by incorporating both multimodal self-supervision and supervision.

\item To demonstrate the applicability of our BMQA, we also establish a new low-light image quality database, namely Dark-4K, which contains only a single image modality. Dark-4K is used to verify the applicability and generalization performance under the unimodal assessment scenarios. Experiments show that this hybrid learning paradigm ensures that BMQA achieves state-of-the-art performance on both the MLIQ and Dark-4K databases.
\end{itemize}

\section{Related Work}\label{sec:related_work}
In this section, we provide a brief review of recent BIQA methods in various aspects, covering hand-crafted BIQAs (\textit{e.g.}, distortion-specific and general-purpose), and deep-learned BIQAs (\textit{e.g.}, supervised learning-based and unsupervised learning-based). In addition, we also review the exploration of multimodality-based quality assessment in user-perceived quality of experience (QoE).


\subsection{Unimodal Blind Image Quality Assessment}

\subsubsection{Hand-crafted BIQA Methods}\label{subsec:traditional_method} 
Hand-crafted BIQAs \cite{wang2021quality} are typically based on the extraction of features from expert and engineering experience. Since hand-crafted BIQAs require less deployment environment (\textit{e.g.}, database size, hardware platform, computing power, \textit{etc}.), they are highly achievable and easy to  deploy. Existing hand-crafted BIQAs can be divided into 1) distortion-specified and 2) general-purposed ones.

\noindent\textbf{Distortion-specific}. 
Distortion-specific BIQAs  measure image quality by considering both degradation manners and distortion types of a particular application, as shown in Fig. \ref{fig:illustration_mliq_database}. For example, some preliminary BIQA models have been specially developed for low-light images, such as brightness-guided \cite{xiang2020blind}, colorfulness-inspired \cite{wang2021blind}, visibility-induced \cite{wang2022low}, and comparative learning-based \cite{wang2022super}.

Additionally, there are other promising distortion-specific BIQAs, including screen content \cite{wang2021quality}, high dynamic range \cite{jiang2021blind}, stereoscopic omnidirectional   \cite{xu2022viewport},  aesthetic \cite{deng2017image}, light-field \cite{tian2020light}, and underwater \cite{zheng2022uif}. Due to the length of the article, we will not discuss other types of distortion-specific BIQAs in this article.

\noindent\textbf{General-purpose BIQAs}. 
General-purpose methods, on the other hand, utilize common quality-aware features to quantify image distortion. For example, natural scene statistics (NSS)-based BIQAs are based on the assumption that high-fidelity images obey some kind of prior statistical characteristics \cite{zhang2015feature}, which are altered by quality degradation in frequency domain \cite{saad2012blind} and spatial domain \cite{mittal2012no}. HVS-guided BIQAs are based on the understanding that the ultimate recipients of visual signals are humans, and it is significant to exploit perception characteristics, such as free-energy principle  \cite{zhai2012psychovisual} and visual sensitivity  \cite{liu2018no}.

\subsubsection{Deep-learned BIQA Methods}
Deep-learned BIQAs \cite{wang2022super} directly learn quality features from distorted images in an end-to-end manner. Unlike hand-crafted BIQAs, these methods automatically optimize quality forecasting models which have shown promising performance. Deep-learned BIQAs can be usually divided into 1) supervised learning-based and 2) unsupervised learning-based BIQA methods. 

\noindent\textbf{Supervised Learning-based}.
Supervised learning-BIQAs \cite{ying2020patches} focus on solving the problem of insufficient training samples.  For example, sample-based BIQAs \cite{kang2014convolutional,bosse2017deep,ke2021musiq} are mainly based on expanding the capacity of training samples by utilizing patch-level quality features to predict an image-level score.  
Constraint-based BIQAs \cite{kim2019deep,zhang2020blind} optimize multiple loss functions simultaneously in supervised learning, such as employing multiple losses for multi-scale supervision \cite{wu2020end}, introducing new normalization embeddings into the objective function \cite{li2020norm},  using additional constraints to adjust initialization parameters \cite{zhu2020metaiqa}, and incorporating constraints learned from other databases \cite{ma2021remember, zhang2021uncertainty, zhang2022continual}.

\noindent\textbf{Unsupervised Learning-based}.
Unsupervised learning-BIQAs extract latent features without relying on ground-truth MOS labels.  For example, metric-based BIQAs \cite{madhusudana2022image} employ widely-used distance measurements (\textit{e.g.}, cosine similarity,  Wasserstein distance, \textit{etc}.) to extract latent embedding features. Domain-based BIQAs \cite{liu2019exploiting} commonly design domain alignment constraints, and measure the quality difference between each sample in a source domain based on the error metric defined in a target domain. However, domain-based BIQAs often require strict assumptions, making it challenging to meet the model requirements when the distortion type of a testing image is unknown.

\subsection{Multimodal Quality Assessment}
Multimodal quality assessment methods are still in their early stage, which primarily focusing on video and audio information \cite{min2020study}. For example, video-audio features can be represented and combined to predict the QoE  score.  Current explorations mainly consider the feature combination of spatial perception \cite{cao2021deep}, temporal perception \cite{ying2022telepresence}, and spatio-temporal perception \cite{min2020study}.

In addition, video and audio distortions do not degrade each other (\textit{i.e.}, video degradation does not cause audio degradation, and vice versa).  The QoE scores of video-audio are first measured independently, and then a combination rule is further designed to predict the joint quality.  The combination rule mainly considers addition \cite{hands2004basic}, multiplication \cite{winkler2006perceived}, and weighted Minkowski \cite{martinez2018combining}. 
Due to the neglect of the interaction between cross-media information, it is difficult to accurately estimate the joint QoE by simply combining individual  quality scores.

\subsection{Motivation}
As discussed in \cite{min2020study,pinson2012influence}, there are some multimodal quality databases established based on video-audio pairs. However, very few databases have been established containing the image-text subjective labels. The differences between existing video-audio quality indicators and our image-text method are apparent: 
\begin{itemize}
	\item[1)] The former aims to predict the joint quality of videos and  corresponding audios, while the latter aims to predict the visual quality of images.  Our BMQA focuses on studying the impact of auxiliary modalities on forecasting image quality.

    \item[2)] Existing video-audio quality methods rarely consider deep multimodal learning, including multimodal feature extraction, alignment, fusion,  and more. Our BMQA investigates various cross-media learning modules, from deep quality representation to quality fusion prediction.

    \item[3)] The audio modality is usually unavailable in many quality assessment applications, and existing video-audio quality methods become  inapplicable. Conversely, the text modality can be easily generated using  existing language models \cite{liu2023pre}, and the lack of text is not a problem in our BMQA.
\end{itemize}

It is worth noting that humans are better at measuring image quality by semantic description rather than quantitative score \cite{yang2022fine}, and hence text can be a very valuable modality in the modeling of BIQA. Moreover, the text modality used in this paper can be directly generate by many popular image captioning methods \cite{xu2015show} and recent large language models (LLMs) \cite{liu2023pre}.

\begin{figure}[!t]
\centering
\includegraphics[width=0.48\textwidth]{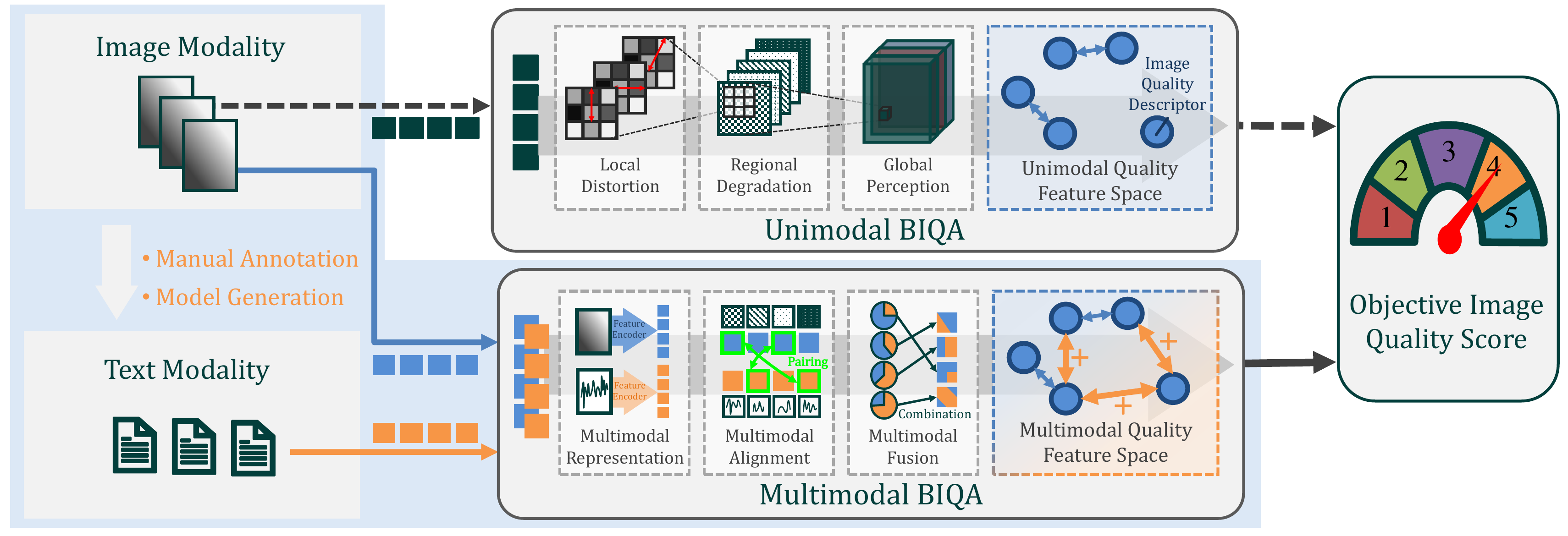}\\
\caption{\textbf{Relationship between unimodal and multimodal BIQAs}. Humans are better at perceiving image quality through semantic descriptions rather than quantitative values, which reveals that text description is a very useful modality for BIQA  modeling.}
\label{fig:unimodal_multimodal}
\end{figure}

Furthermore, our investigation focuses on how additional modality information (\textit{e.g.}, semantic visual understanding \cite{wade2013visual}), if present, affects the modeling of BIQA. With our MLIQ database, we can pre-train a QSD-induced captioning model, allowing our BMQA scheme to seamlessly integrate into any existing unimodal BIQA task.  The relationship between unimodal BIQAs and our BMQA is illustrated in Fig. \ref{fig:unimodal_multimodal}.

Finally, our BMQA provides a new and promising research perspective. On one hand, the homogeneity of multimodal data suggests that training information can be supplementary or shared, which facilitates the learning of highly descriptive quality features. On the other, the heterogeneity of multimodal data can expand the breadth and depth of training information, which potentially improves the forecasting performance. To this end, it is highly desirable to develop a specialized multimodal quality indicator for low-light images.

\section{Multimodal Low-light Image Quality (MLIQ) Database Construction}\label{sec:database_constrution}
\begin{figure*}[!t]
\centering
\begin{tabular}{p{5.3cm}<{\centering} p{5.3cm}<{\centering} p{5.3cm}<{\centering} }
\includegraphics[width=1.06\linewidth, height=.6\linewidth]{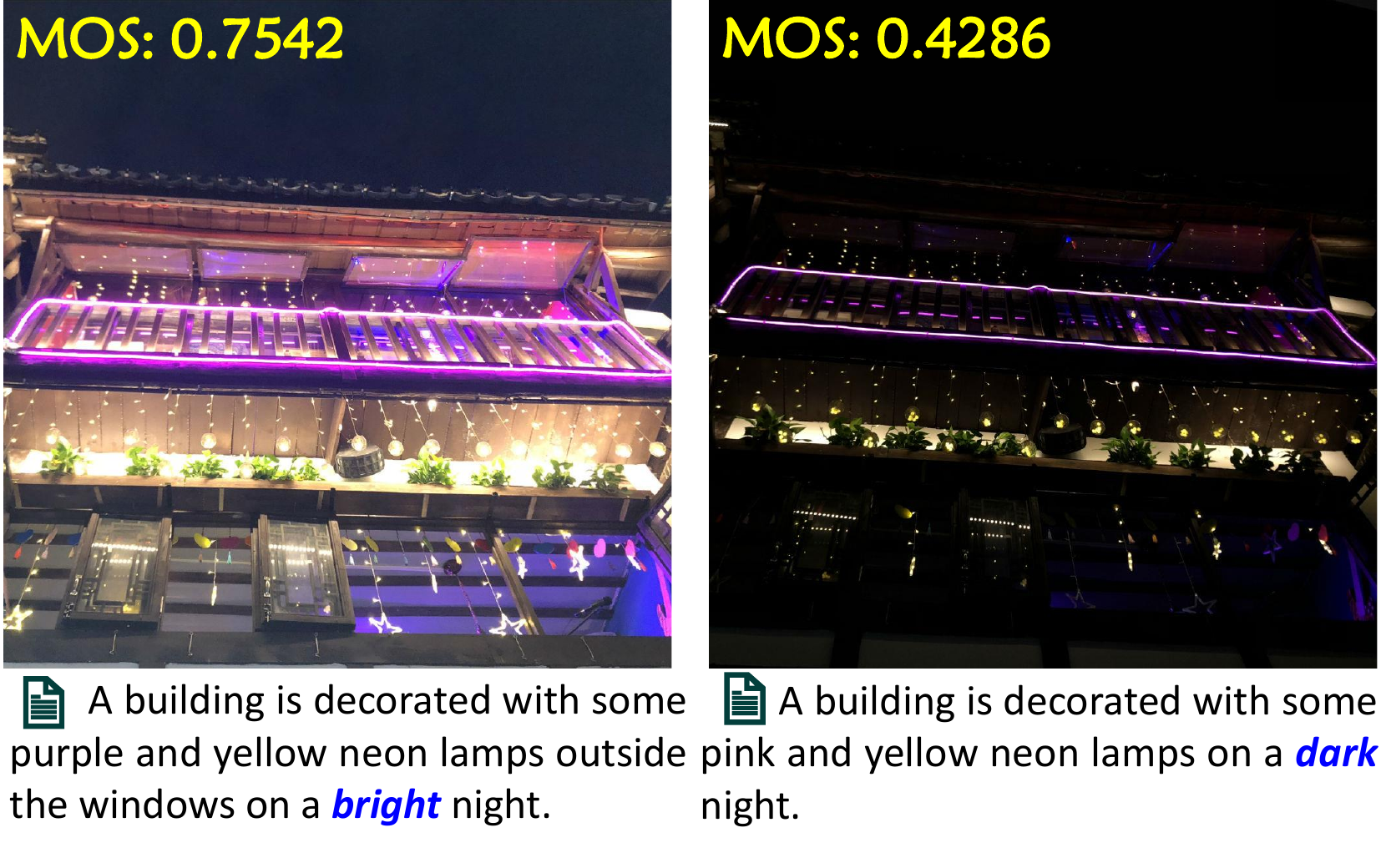} 
&\includegraphics[width=1.06\linewidth, height=.6\linewidth]{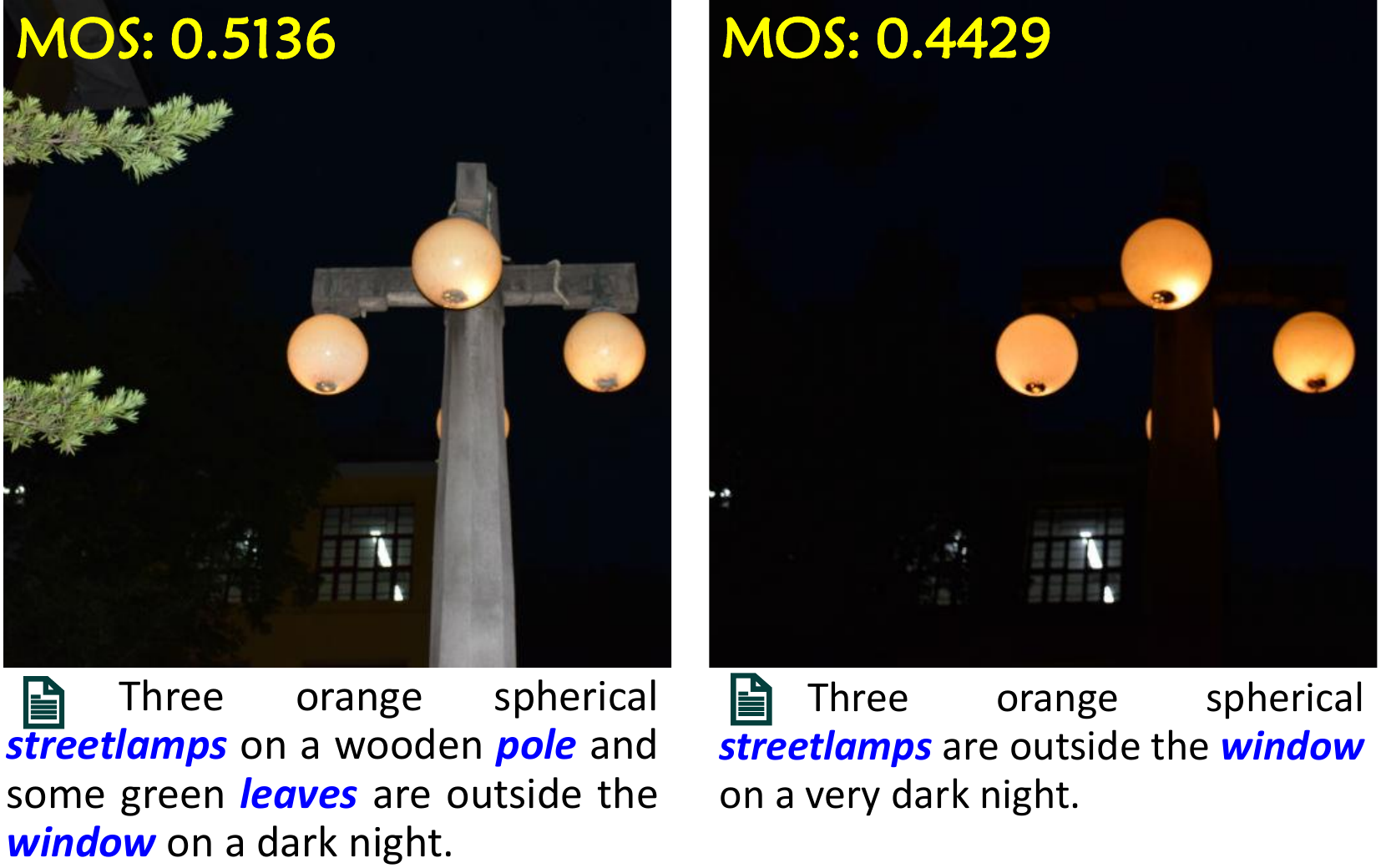} 
&\includegraphics[width=1.06\linewidth, height=.6\linewidth]{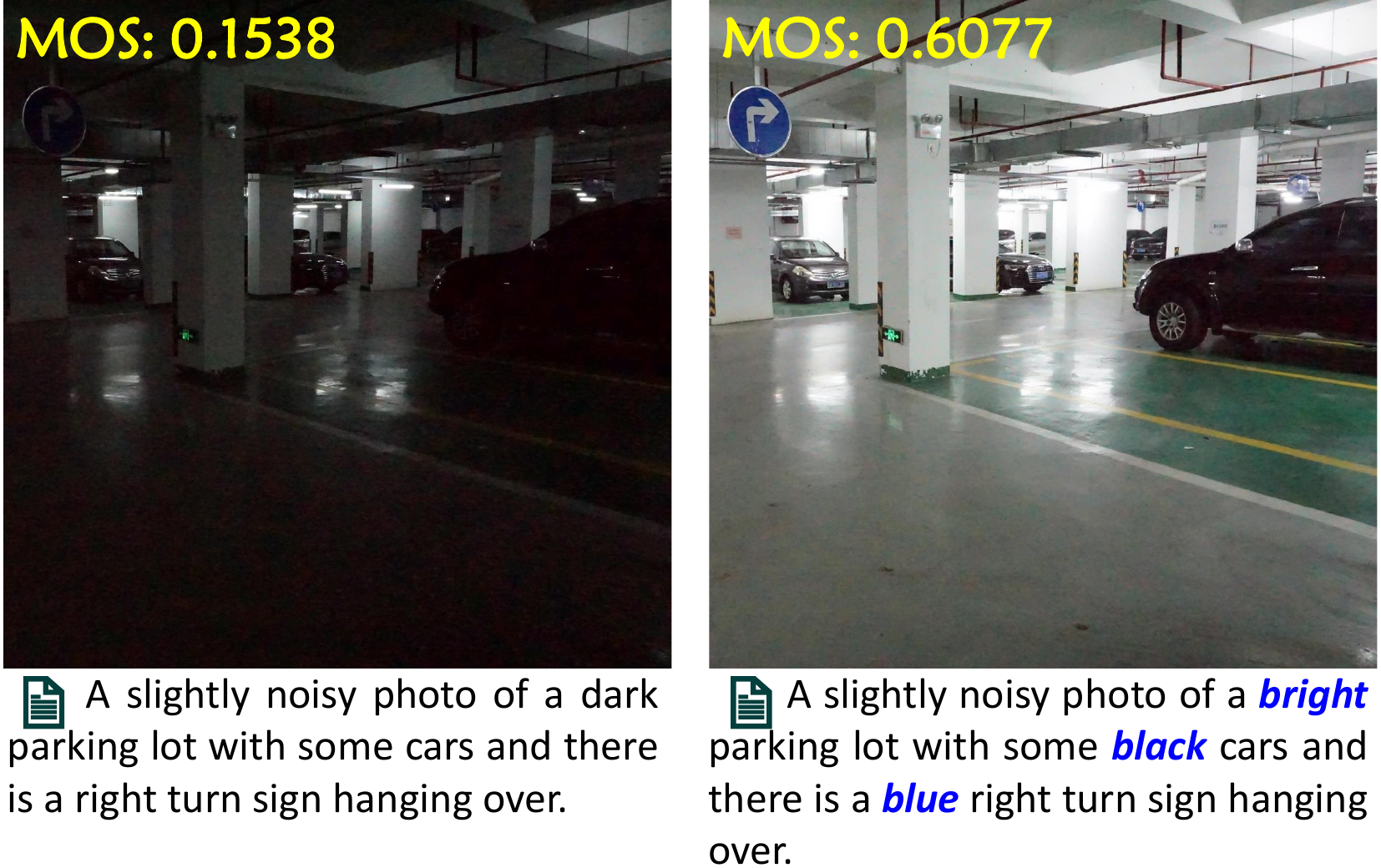}  \\ 
(a) {\scriptsize Luminance} & (b) {\scriptsize Content} & (c) {\scriptsize Color}\\  	
\includegraphics[width=1.06\linewidth, height=.6\linewidth]{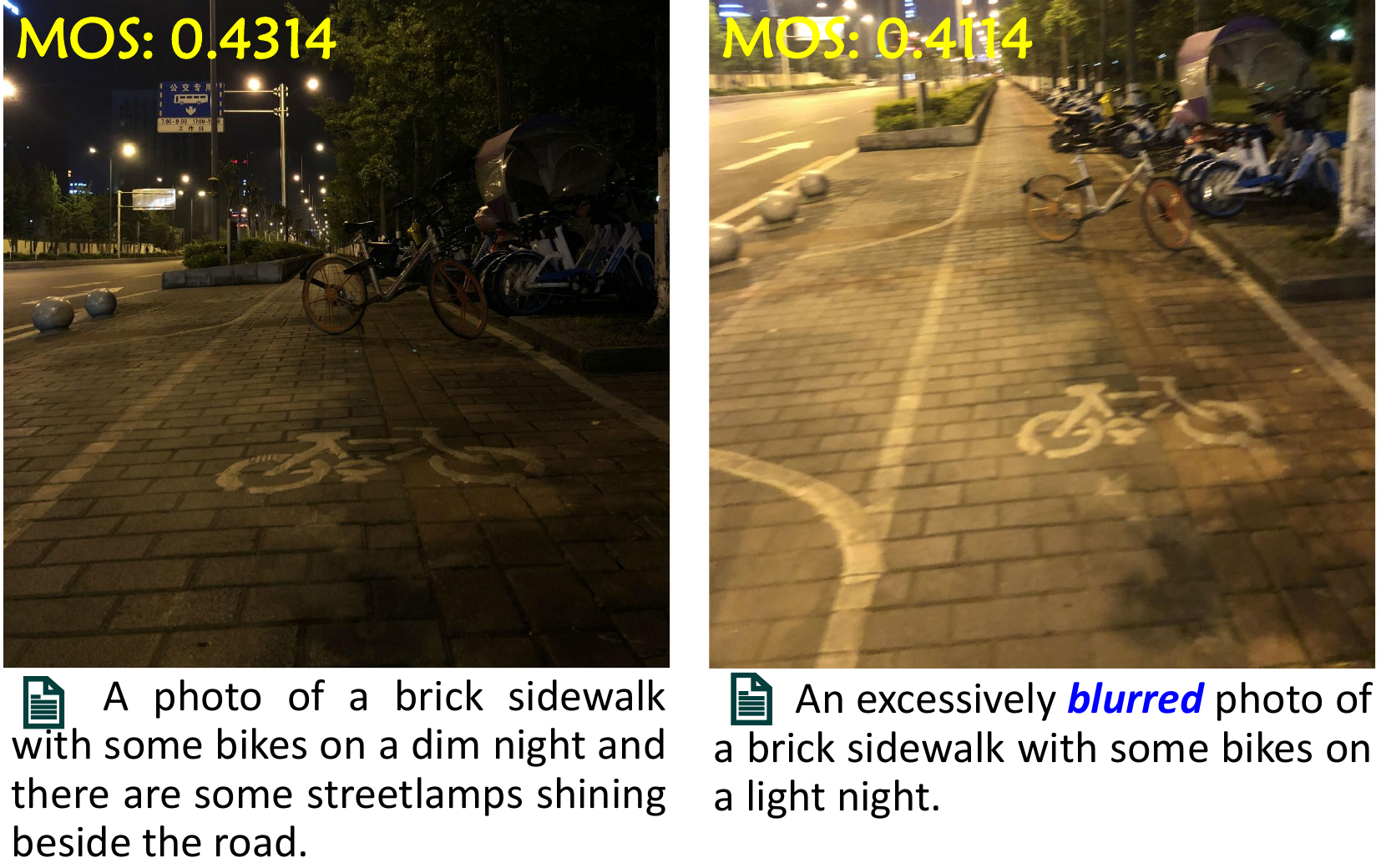} 
&\includegraphics[width=1.06\linewidth, height=.6\linewidth]{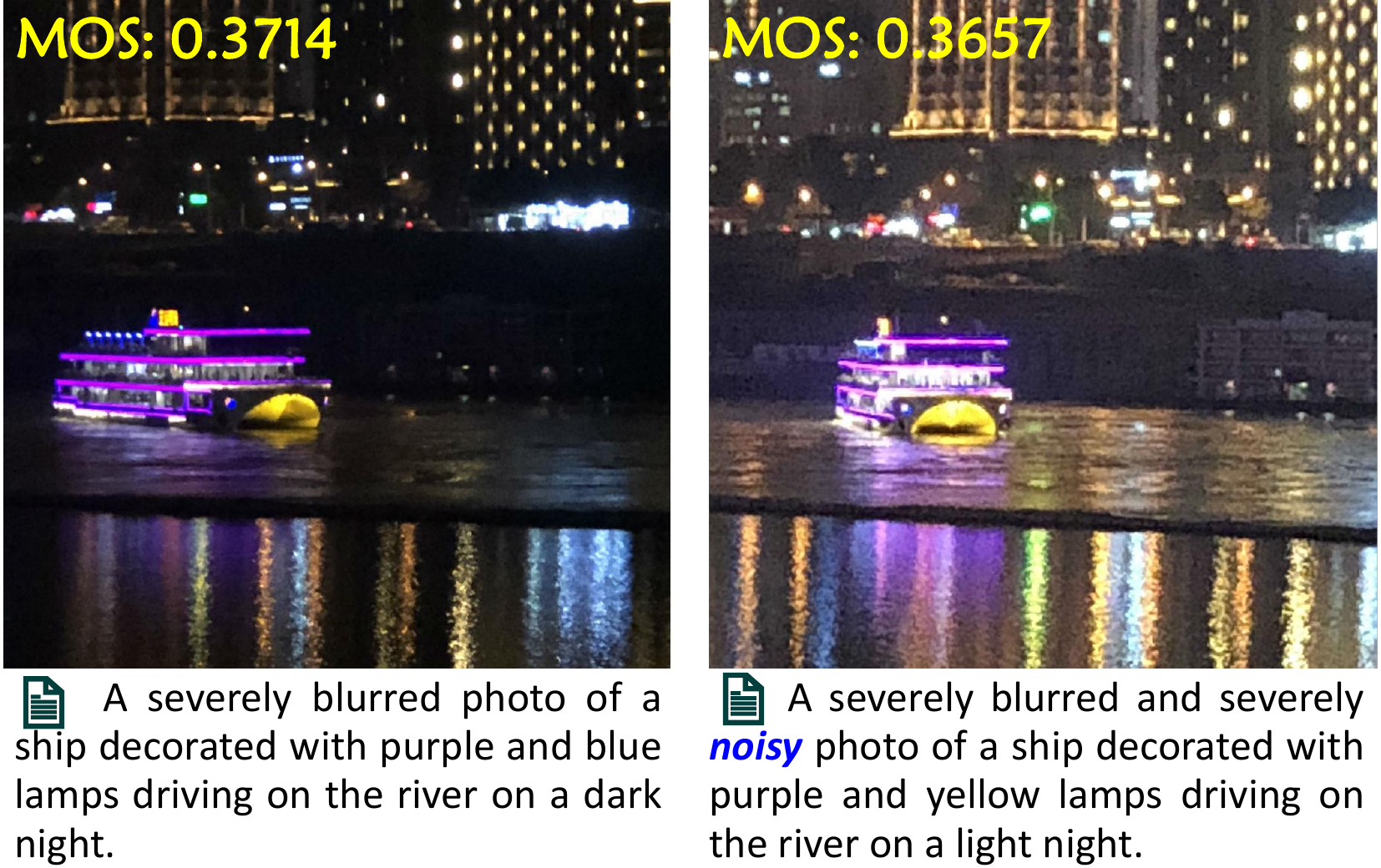} 
&\includegraphics[width=1.06\linewidth, height=.6\linewidth]{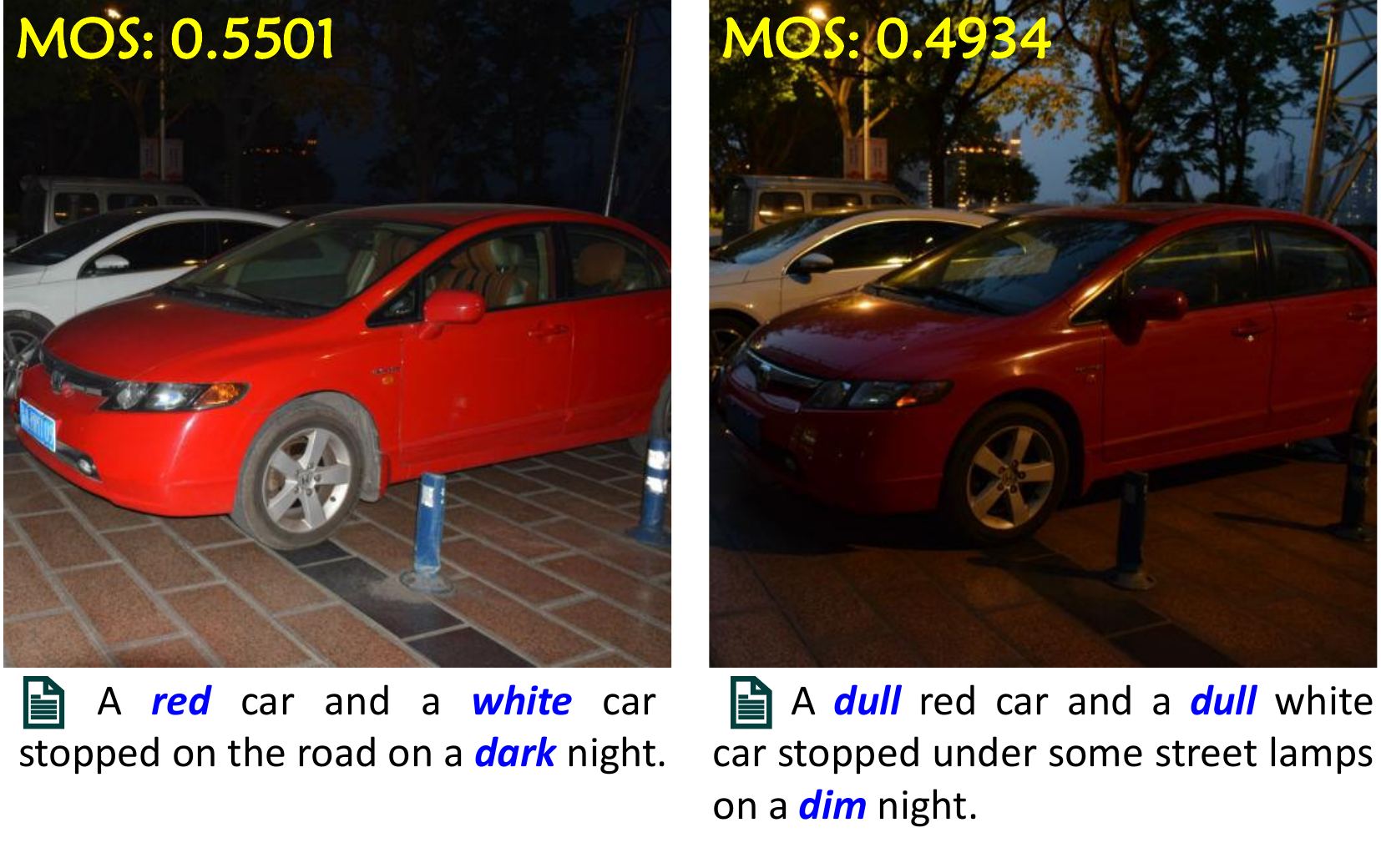}  \\  
(d) {\scriptsize Blur} & (e) {\scriptsize Noise} & (f) {\scriptsize Saturation}\\  
\end{tabular}   
\caption{\textbf{Examples of some Image-Text-MOS pairs on our MLIQ database}. We provide some representative examples for analysis and discussion of the text attributes, including luminance, content, color, blurry, noise, and saturation. Please zoom in the electronic version for better details.}
\label{fig:visualization_examples}
\vspace{-0.3cm}
\end{figure*}

In this section, we describe how to construct our MLIQ database. The established database contains RGB images with the subjective quality scores and QSD-based texts  as shown in Fig. \ref{fig:visualization_examples}. 

\subsection{Multimodality Construction}\label{subsec:multi_modality_construction}
\subsubsection{MOS-based Image Modality}
\label{subsubsec:image_modalilty}
Natural Night-time Image Database (NNID) \cite{xiang2020blind} is the latest publicly available no-reference low-light image database. It contains 448 different visual scenes with a total of 2240 samples, covering daily life recording, intelligent transportation, surveillance, city light show, aerial photography, and many other application scenarios. The characteristics of the NNID database are very suitable for our experimental requirement. To facilitate the acquisition of low-light image data for MLIQ, we adopt it as part of our benchmark image source.

Furthermore, we adopt another 1360 low-light images captured by two new  devices, {\textit{Canon EOS 6D}} and \textit{Huawei Mate 30 Pro}, to expand the coverage of weak-illumination scenes. MLIQ consists of a total of 3600 low-light image samples. These low-light samples are captured in real-world environments (\textit{e.g.},  indoors and outdoors) with a total of five different mobile devices. One device captures a visual scene with five different settings. These five settings are allowed to be different for different scenarios. The resolution of each low-light sample ranges from {512$\times$512 to 2048$\times$2048}. Therefore, MLIQ is the largest no-reference low-light database, covering various scenes, large volumes, complex noise, diverse devices, and authentic distortion.

To obtain the MOS label for each low-light image on MLIQ, we have conducted a subjective experiment. Following ITU-R BT.500-14 \cite{bt2020methodologies} and NNID \cite{xiang2020blind}, we take a single stimulus and build a graphical user interface to perform the experiments.  \textit{Skyworth 28U1} is used, and the viewing distance is approximately three times of the image height.  
We have invited 26 participants including 16 females and 10 males (between the ages of 18 and 35). The participants are asked to score the image quality based on 11 discrete quality scores ranging from 0 to 1 with a step length of 0.1. 
The rated score will finally correspond to five quality levels, including $[0,0.1]$ for \textit{bad}, $[0.2,0.3]$ for \textit{poor}, $[0.4,0.5]$ for \textit{fair}, $[0.6,0.7]$ for \textit{good}, and $[0.8,1.0]$ for \textit{excellent}.

\begin{figure}[!t]
\raggedright
\begin{tabular}{p{2.5cm}<{\centering} p{2.5cm}<{\centering} p{2.5cm}<{\centering}}
\includegraphics[width=1.1\linewidth]{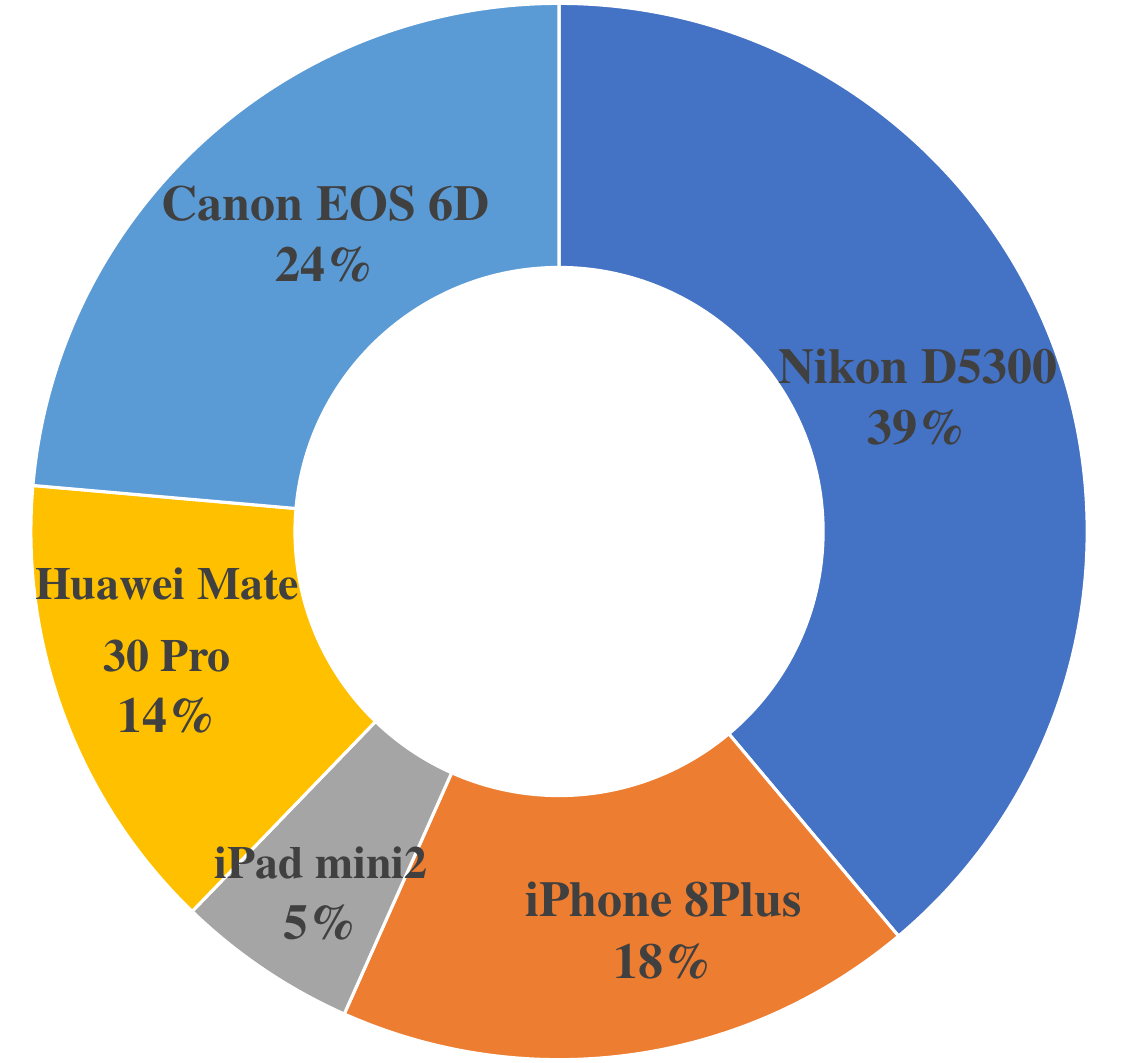}&\includegraphics[width=1.1\linewidth]{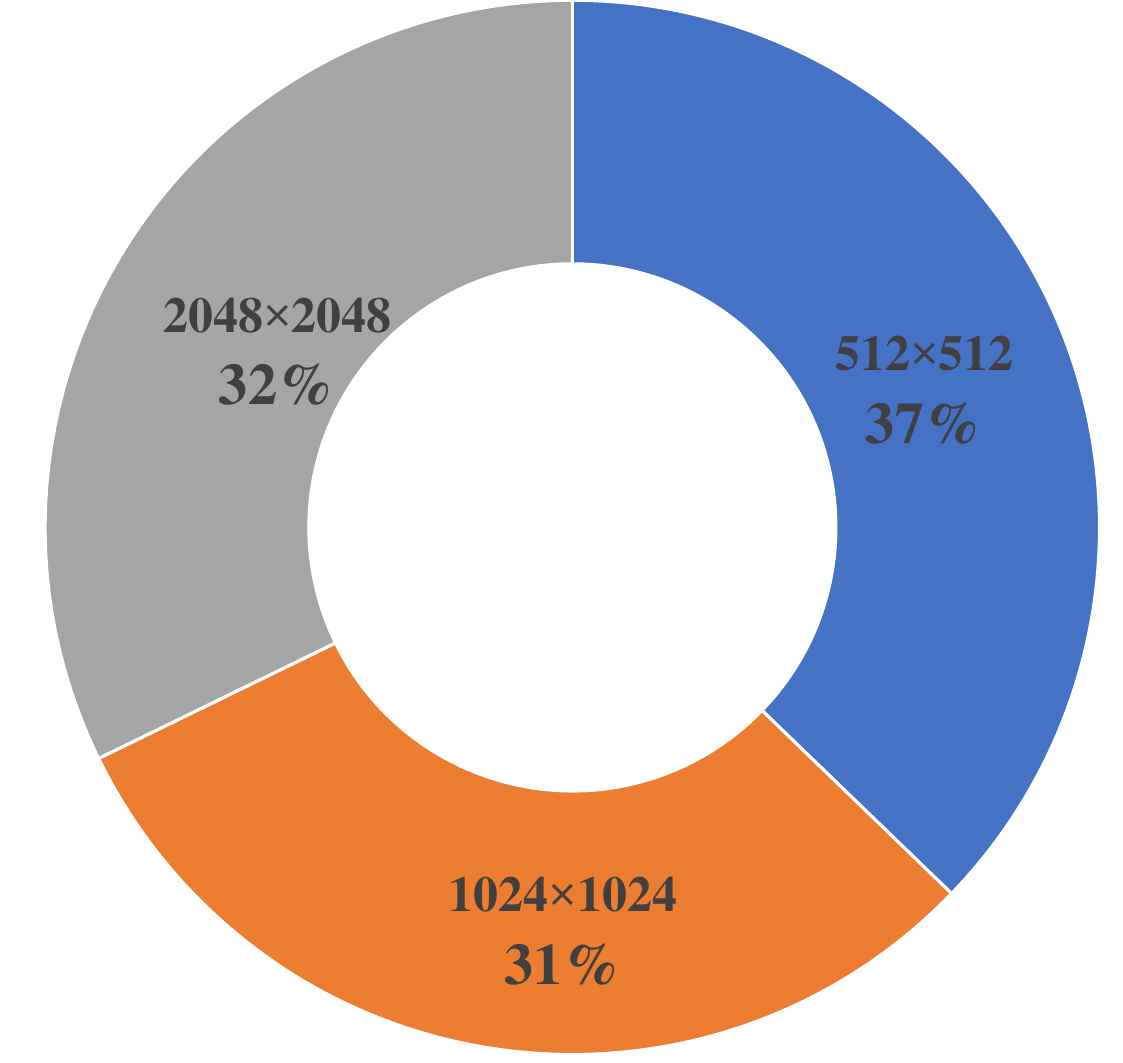}&\includegraphics[width=1.1\linewidth]{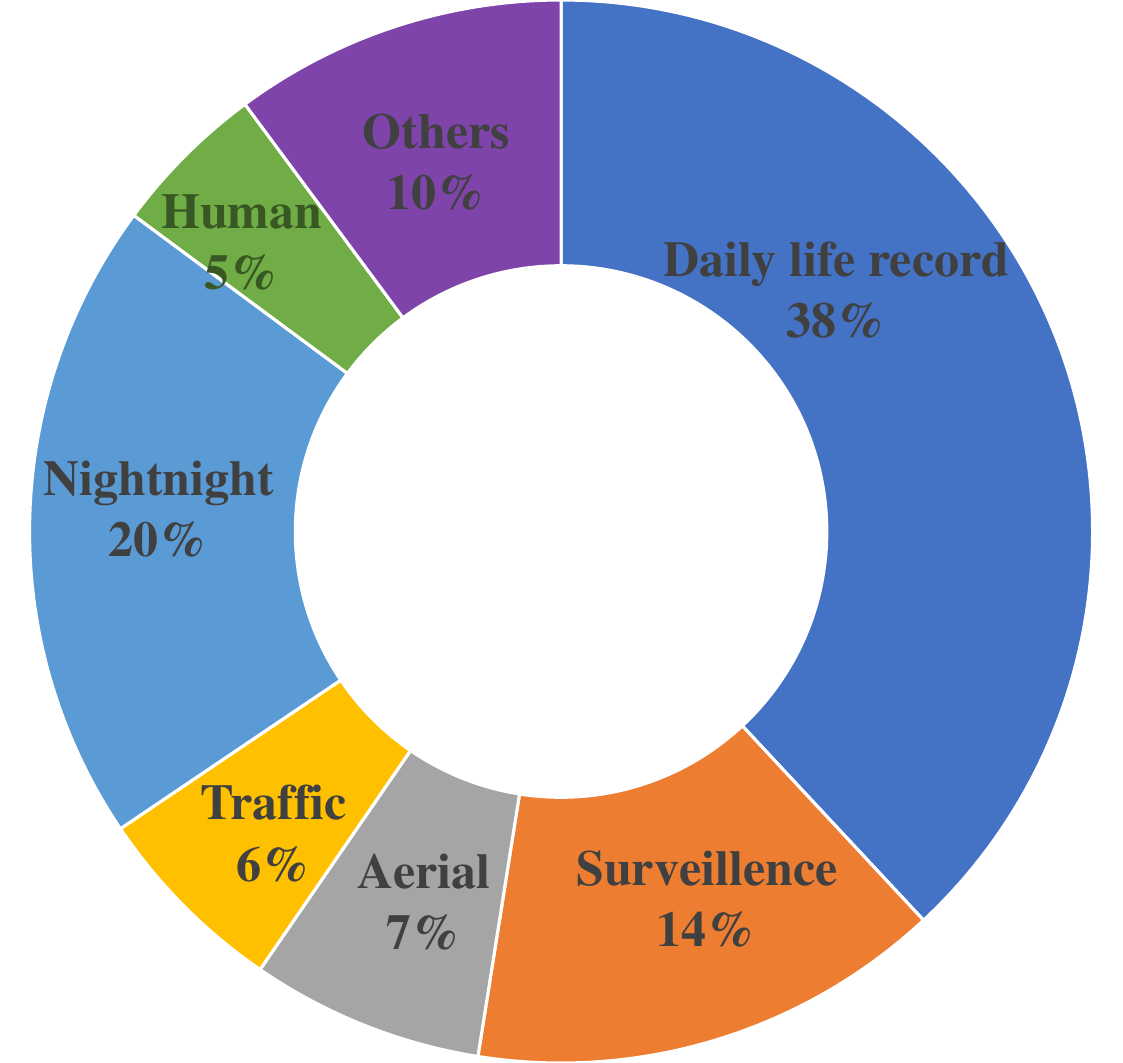}\\  
(a) {\scriptsize Device} & (b) {\scriptsize Resolution} & (c) {\scriptsize Application}\\  	
\end{tabular}    
\begin{tabular}{p{3.75cm}<{\centering} p{3.75cm}<{\centering}}
\includegraphics[width=1.1\linewidth]{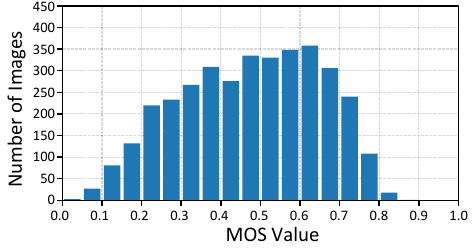}&\includegraphics[width=1.1\linewidth]{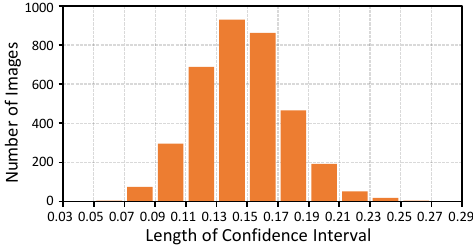} \\  
(d) {\scriptsize MOS distribution} & (e) {\scriptsize Confidence interval} \\  	
\end{tabular}  
\caption{\textbf{A statistical analysis of the proposed multimodal MLIQ database}: (a) shooting device, (b) image resolution,  (c) application scenario, (d) MOS distribution, and (e) confidence interval.}
\label{fig:mliq_statistics_img}
\end{figure}

A statistical analysis of our MLIQ is illustrated in Fig. \ref{fig:mliq_statistics_img}. Figs.~\ref{fig:mliq_statistics_img} (a)-(c) provide the overall statistical data of shooting device, image resolution, and content application scenario, respectively. Fig.~\ref{fig:mliq_statistics_img} (d) reports the histogram distribution of MOS values.  As seen, MOS values span the entire quantified range of visual quality with sufficient and fairly uniform samples at each level. This shows that our MLIQ database covers the entire range of visual quality (from poor to excellent), and also exhibits a good separation of the perceptual quality. In addition, Fig.~\ref{fig:mliq_statistics_img} (e) reports the $95\%$ confidence intervals obtained from the mean and standard deviation of the rating score for each image as the consistency evaluation, where the confidence interval is mainly distributed between $0.11$ and $0.17$. It indicates that all observers have reached a high agreement on the perceptual quality of low-light images. Therefore, the proposed MLIQ database is used as a ground-truth for the performance evaluation of objective quality indicators.

\subsubsection{QSD-based Text Modality}
\label{subSec:text_modality}
A simple and effective way to represent human visual perception is to ask subjects to describe and record their semantic visual understanding \cite{wade2013visual}. The text description can provide additional quality assessment clues: On one hand, free verbal descriptions of visual understanding provide potentially the richest QSD source, since language is the most flexible means of human communication. On the other, verbal descriptions help to avoid personal information bias, as verbal descriptions are far less likely to be inconsistent than consistent \cite{hanjalic2005affective}.

By synthesizing some previous work, we design two text description principles based on the perception mechanism of the HVS. It can exhibit feed-forward visual information extraction and aggregation from the retina (\textit{i.e.}, \textit{intuitive visual perception}) to the primary visual cortex (\textit{i.e.}, \textit{empirical visual perception}) \cite{yang2022inferring}. These principles are used to guide annotators in generating their verbal descriptions.

\noindent\textbf{Intuitive Visual Perception.}
This principle is inspired by previous physiological and psychological experiments on the HVS, including saliency detection and just noticeable difference. It is closely related to early vision and focuses on the relationship between optical stimuli from visual observation and the HVS experience \cite{wade2013visual}. Intuitive vision mainly focuses on some basic data characteristics, covering overall brightness, color status, texture, salient objects, \textit{etc}. 
For instance, the verbal description in Fig. \ref{fig:visualization_examples} (c) contains the quality attributes, such as the luminance information `\textit{bright}', the color information `\textit{black}' and `\textit{blue}', and the observed object information `\textit{car}' and `\textit{sign}'.

\noindent\textbf{Empirical Visual Perception.}
This principle is inspired by modern theoretical studies in visual theory that embraces empiricism as the dominant paradigm of human perception construction, such as Gregory's theory \cite{gregory1980perceptions}. These studies demonstrate that while part of what we perceive comes from objects in front of us, another part (and possibly a larger part) always comes from our brains. 
The empiricism principle is closely related to late vision and focuses on exploring how the viewpoints of observers are involved in quality cognition. Empirical vision mainly involves some common real-world knowledge as well as empirical visual understanding, and highlights the possible real-life scenarios of low-light observations \cite{gordon2004theories}. For instance, subjects use `\textit{driving}' rather than `\textit{sitting}' for `\textit{ship}' as shown in Fig. \ref{fig:visualization_examples} (e).

However, there are several challenges associated with verbal description. In this article, we will consider three of them in obtaining the text labels:
\begin{itemize} 
\item \textbf{Subjectivity}. Verbal description of image quality perception is hindered by inherent subjectivity in individual preferences, making it difficult to establish a standardized and universally applicable vocabulary for describing image quality.

\item \textbf{Variability}. Verbal descriptions exhibit significant variability.  Individuals perceive and interpret the same test image differently based on personal experiences, cultural backgrounds, aesthetic preferences, and physiological structures. In other words, visual appeal may vary among individuals.

\item \textbf{Expressiveness}. Human language has limited capacity to express visual attributes and qualities. Describing with words often falls short in capturing the richness and complexity of image content. For instance, it can be challenging to articulate subtle differences in color tones, texture, or lighting conditions accurately. 
\end{itemize}

To address the above three challenges, we attempt to develop a tractable verbally description paradigm. The quality-based sentence should include a set of perceptual attributes such as color accuracy, noise, sharpness, and overall image quality. In the experiments, each subject is asked to provide a meaningful and concise verbal description for one low-light image. For the dictation of each image content, the following requirements need to be met: 
\begin{itemize}
\item For images with salient objects (\textit{e.g.}, the ship in Fig. \ref{fig:visualization_examples} (e)), trying to describe all important objects in the image content. For images with salient scenes (\textit{e.g.}, the parking scene in Fig. \ref{fig:visualization_examples} (c)), trying to describe the overall environment. For images without any salient content (\textit{e.g.}, the building with many small objects in Fig. \ref{fig:visualization_examples} (a)), trying to describe attractive content part, including objects and scenes.	
	
\item Trying to describe the overall brightness by using the relevant lighting  features, such as  `\textit{bright}', `\textit{light}', `\textit{dim}', or `\textit{dark}'. 
	
\item Trying to describe the main attributes of each object, such as color, brightness, texture (\textit{e.g.}, `\textit{wooden}' pole in Fig. \ref{fig:visualization_examples} (b) and `\textit{brick}' sidewalk in Fig. \ref{fig:visualization_examples} (d)), \textit{etc}.
	
\item Trying to describe the quality cognition by using the relevant color  features, such as `\textit{colorful}', `\textit{vivid}', `\textit{blurred}', `\textit{noisy}', \textit{etc}. 
\vspace{-0.25cm}
\end{itemize}

In the experiments, 14 male and 10 female trained participants are invited to provide a meaningful and concise verbal description for each low-light image. Finally, we collect 3600 text descriptions, where the length of the text labels ranges from 6 to 45 words.

\subsection{Database Analysis}
In this section, we analyze correlations between images, texts, and quality scores (\textit{i.e.}, MOS) on the MLIQ database. Specifically, the image modality represents a visual stimulus, and the corresponding text information represents a subjective quality cognition and understanding of the associated  image modality. Based on the statistical analysis of MLIQ, we attempt to capture the underlying connection between visual signals and verbal descriptions. We conduct the statistical analysis based on brightness, content and  color, and then discuss other factors affecting quality perception on low-light images.

\subsubsection{Luminance}
Low-light images mainly suffer from insufficient or uneven brightness. The quality level of low-light images is sensitively dependent on visual brightness perception. The text information contains quality clues (\textit{i.e.}, keywords) that describe the luminance status, which can effectively provide supplementary information. Therefore, it is significantly meaningful to explore the relationship between brightness and quality.

\begin{figure}[!t]
\centering
\begin{tabular}{p{2.5cm}<{\centering} p{2.5cm}<{\centering} p{2.5cm}<{\centering} }
\includegraphics[width=1.15\linewidth, height=.9\linewidth]{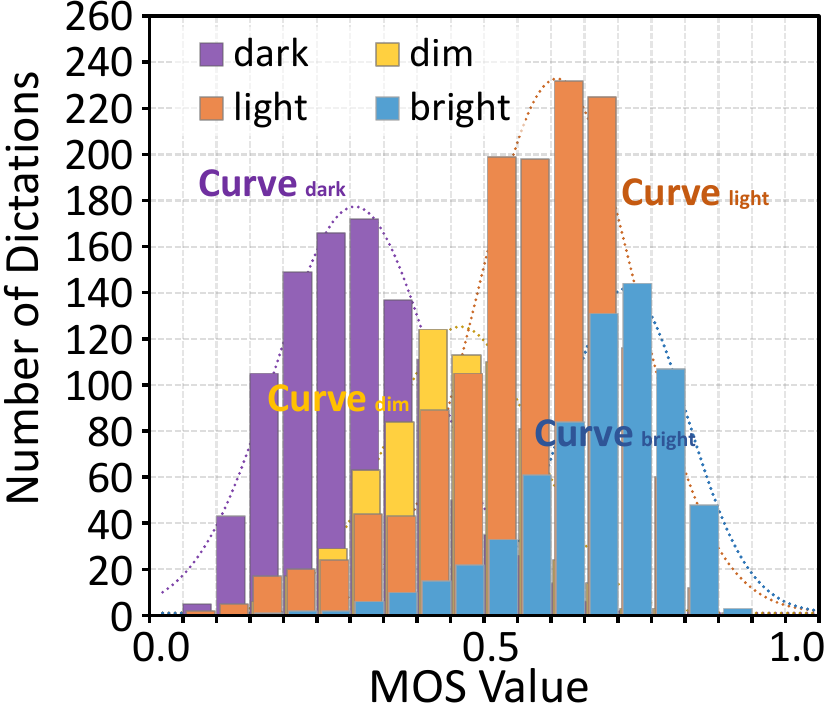} 
&\includegraphics[width=1.15\linewidth, height=.9\linewidth]{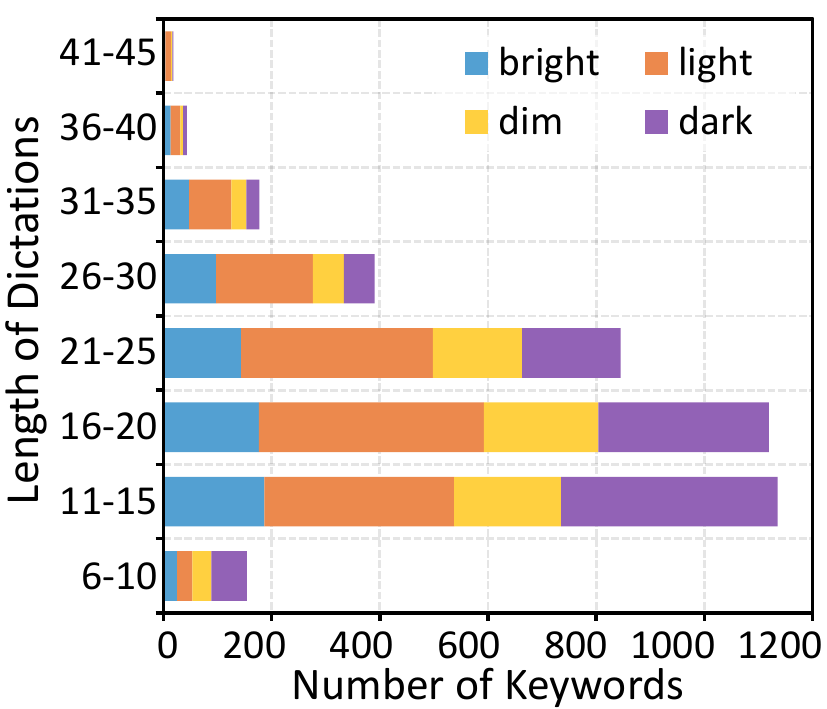} 
&\includegraphics[width=1.15\linewidth, height=.9\linewidth]{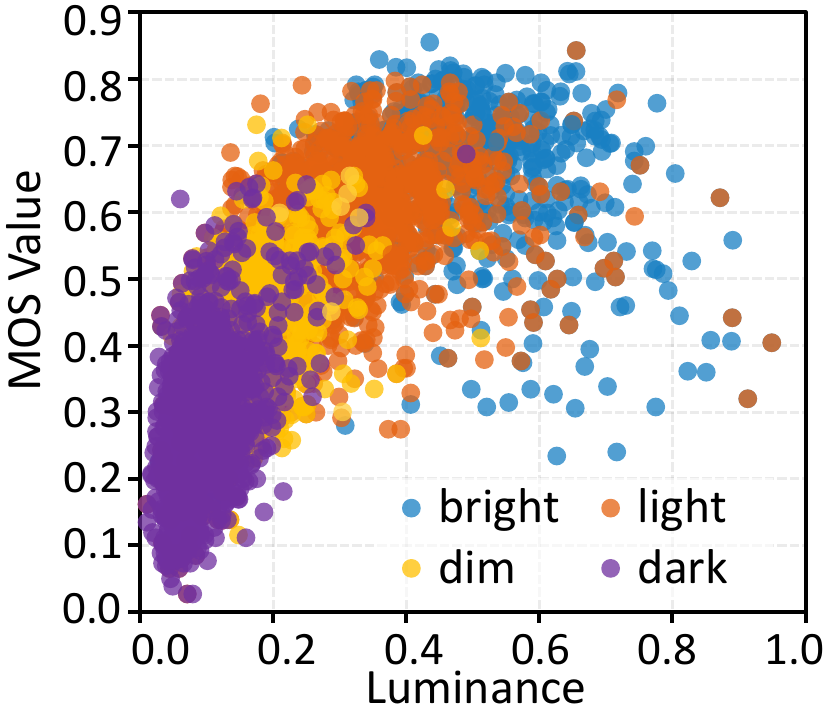}  \\  
(a) {\scriptsize Gaussian fitting} & (b) {\scriptsize Keyword length} & (c) {\scriptsize Subjective quality}\\  	
\end{tabular}     
\vspace{2pt}
\caption{\textbf{Statistics of text quality feature based on luminance}: (a) histogram of the luminance keywords  and Gaussian fitting,  (b) histogram of the length and number of luminance keywords, and (c) scatter distribution of the luminance keywords and MOS values.}
\label{fig:relation_image_text_mos}
\end{figure}

We start by figuring out how QSD-based text represents the image quality from the view of brightness. The luminance quality features, covering `\textit{dark}', `\textit{dim}', `\textit{light}', and `\textit{bright}',  represent the  illumination condition. 
We calculate the histogram of the luminance feature on the entire database, as shown in Fig.~\ref{fig:relation_image_text_mos} (a). As seen, the MOS value corresponding to `\textit{dark}', `\textit{dim}', `\textit{light}', and `\textit{bright}' is concentrated around $0.2$ to $0.4$, $0.4$ to $0.5$, $0.5$ to $0.6$, and $0.6$ to $0.8$, respectively.
Intuitively, the histogram of each luminance feature should obey an independent Gaussian distribution. Therefore, we adopt a Gaussian function to fit the histogram of `\textit{dark}', `\textit{dim}', `\textit{light}', and `\textit{bright}', and the Gaussian centers are $0.2797$, $0.4338$, $0.5744$, and $0.6745$, respectively. 
The above observations suggest that the luminance attribute has a strong relationship with the image quality.

Next, we analyze the relationship between images, texts, and quality scores based on  brightness. We report the stacked histogram of the length of verbal at various luminance conditions in Fig.~\ref{fig:relation_image_text_mos} (b). As seen, `\textit{dim}' and `\textit{dark}' represent low luminance, which tend to have shorter verbal lengths. This is consistent with our experience that people often have difficulty describing very dark scenes with long verbal descriptions.

In addition, we calculate the average luminance value for each image as the objective luminance level and report the scatter plot of the corresponding quality score as shown in Fig.~\ref{fig:relation_image_text_mos} (c). We further mark colors for each plot based on the luminance keyword contained in the corresponding text. It can be observed that as the luminance level increases, the quality score generally increases as shown in different scatter colors. It suggests that the luminance feature is an efficient representation of visual quality perception.

\subsubsection{Content}
Due to insufficient exposure, low-light distortions usually result in incomplete or unclear visual quality, further leading to an annoying visual experience. The text  modality contains verbal descriptions of observed objects, which can effectively provide auxiliary information on which objects the visual attention is focused on. Therefore, we explore the relationship between image quality and content.

The reduction of observed objects is often reflected in the reduction of object descriptions in the text modality, as shown in Fig.~ \ref{fig:visualization_examples} (b). Based on this observation, we count the quantity of observed objects and report the corresponding MOS values. Fig. \ref{fig:bars_centent_color_keywords} (a) consists of stacked column charts and scatter plots, covering the number of observed objects ranging from 1 to 5.
For each number of observed objects, the stacked column chart reports the image number at each luminance level, including `\textit{dark}', `\textit{dim}', `\textit{light}', and `\textit{bright}', respectively.

Based on the statistical data of image content, we can draw some interesting conclusions: 
1) The curve in Fig. \ref{fig:bars_centent_color_keywords} (a) shows that the quality score tends to be higher as the quantity of observed objects increases.  This may indicate that images with better visual quality usually contain more identifiable observed objects.
2) The stacked histogram in Fig. \ref{fig:bars_centent_color_keywords} (a) shows that when the luminance levels get lower, the number of observed objects decreases. 
3) The quality score increase caused by the quantity increase of observed objects is small ({$0.0657$} from 1 to 5), which indicates that it is difficult to sensitively reflect visual experience via the quantity of observed objects. One possible reason may be that low-light distortion tends to lose detail rather than salient objects, while the quality score depends more on the salient content itself.

\subsubsection{Color}
Low-light distortion tends to exhibit low color contrast and low saturation. The observed colors may effectively provide useful text information on visual perception responses. Therefore, we investigate the relationship between image quality and observed colors.

\begin{figure}[!t]
\centering
\begin{tabular}{p{2.5cm}<{\centering} p{2.5cm}<{\centering} p{2.5cm}<{\centering} }
\includegraphics[width=1.15\linewidth, height=.9\linewidth]{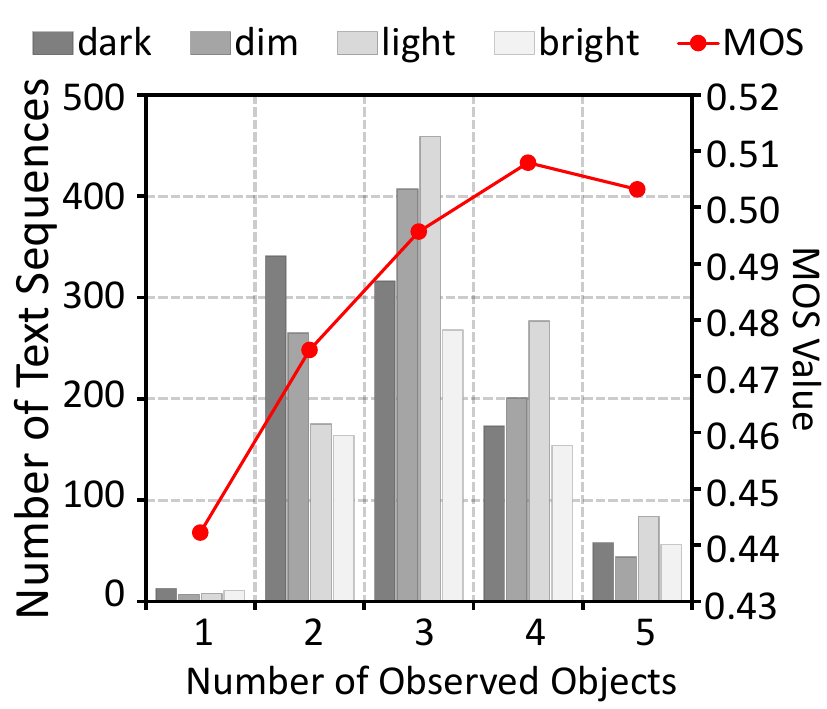} 
&\includegraphics[width=1.15\linewidth, height=.9\linewidth]{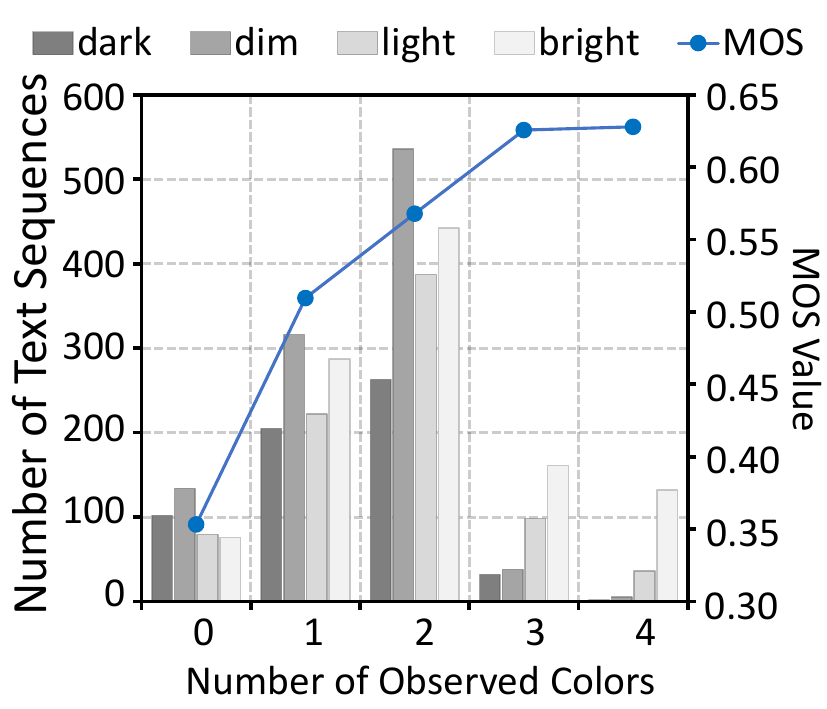} 
&\includegraphics[width=1.15\linewidth, height=.9\linewidth]{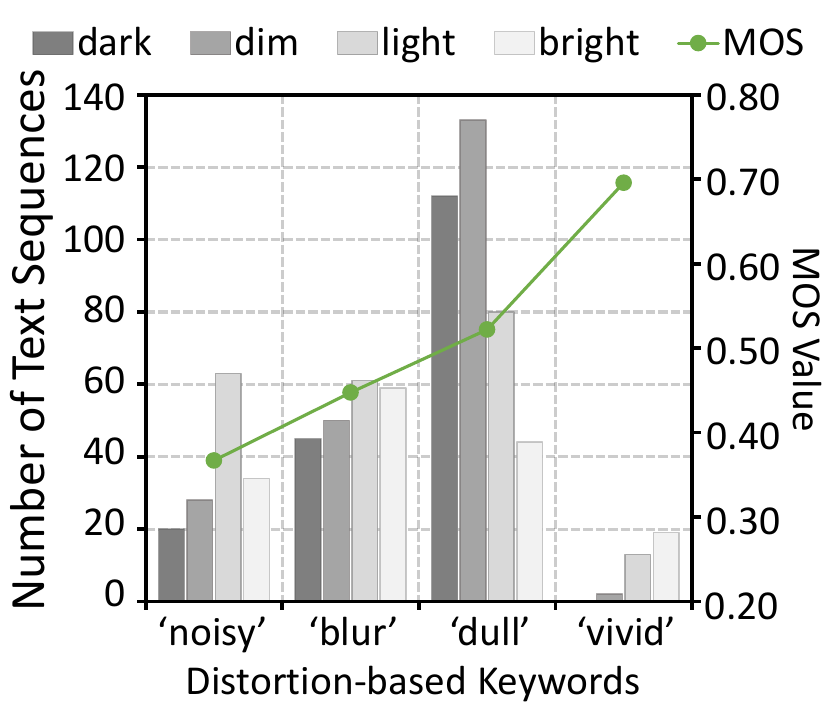}  \\  
(a) {\scriptsize Content} & (b) {\scriptsize Color} & (c) {\scriptsize Keyword}\\  	
\end{tabular} 
\vspace{2pt}   
\caption{\textbf{Statistics of other text quality features}: (a) content, (b) color, and (c) keyword. The number of related images is measured by the main ordinate (placed on the left-hand side), while the average MOS result (dots connected by a solid line) is measured by the secondary ordinate (placed on the right-hand side).}
\label{fig:bars_centent_color_keywords}
\end{figure}

The impairment of observed colors is reflected as the reduction of color descriptions, as shown in Fig.~ \ref{fig:visualization_examples} (c). Inspired by this, we count the number of color features and report the corresponding MOS values. Fig.~\ref{fig:bars_centent_color_keywords} (b) consists of stacked column charts and scatter plots, covering the number of color words ranging from 0 to 4. For each number of color features, the stacked column chart reports the image quantity at each luminance level, including `\textit{dark}', `\textit{dim}', `\textit{light}' and `\textit{bright}', respectively.

Based on the statistical data of image color, we can also draw some interesting conclusions:  1) The curve in Fig.~\ref{fig:bars_centent_color_keywords} (b) shows that the quality score tends to be higher as the number of color features increases. This may indicate that images with a better visual perception experience usually contain more recognizable colors, as shown in Fig.~\ref{fig:visualization_examples} (c). 2) The stacked histogram in Fig.~\ref{fig:bars_centent_color_keywords} (b) shows that when the visual perceptual luminance gets lower, the number of observed colors tends to be lower. 3) The quality score increases caused by the number increase of observed color is large ($0.2747$ from 0 to 4). This may indicate that the quantity of observed color can sensitively represent the quality experience of low-light distortion.

\subsubsection{Other Factors}
Low-light photography is also often affected by many other factors, including blurring, heavy noise and low saturation \cite{chen2021learning}. A low-light image may get blurred by the camera shake if it is set to a long exposure time, as shown in Fig.~ \ref{fig:visualization_examples} (d).  The increase in light sensibility reduces the signal-to-noise ratio while increasing the exposure, as shown in Fig.~ \ref{fig:visualization_examples} (e).  In addition, both underexposure and overexposure significantly affect color saturation, which further affect the visual experience.

Considering that the text label may contain some keywords that directly describe these degradation features, we collect the distortion-based keywords and report the corresponding MOS values. Fig.~\ref{fig:bars_centent_color_keywords} (c) shows stacked column charts and scatter plots, covering the distortion-related keywords of `\textit{blur}', `\textit{noisy}', `\textit{dull}', and `\textit{vivid}'.

Based on the above statistical data, we can draw some interesting conclusions:
1) text features such as `\textit{blur}', `\textit{noisy}', and `\textit{dull}' represent poor visual experience, while `\textit{vivid}' represents good visual experience.	
2) The proportion of `\textit{noisy}' is large in the `\textit{bright}' luminance level, which indicates that noises in low-light images are more easily perceptible. 
3) The proportion of `\textit{blur}' is similar under different luminance levels, which indicates that blur is not closely related to lumination.
4) The proportion of `\textit{dull}' is large in the `\textit{dark}' and `\textit{dim}' luminance levels, while the proportion of `\textit{vivid}' is large in the `\textit{light}' and `\textit{bright}' luminance levels. 
These observations are consistent with the fact that human eyes prefer highly saturated colors, as shown in Fig.~ \ref{fig:visualization_examples} (f).

\section{Proposed Blind Multimodal Quality Assessment (BMQA) Framework}
\begin{figure}[!t]
	\centering
	\includegraphics[width=0.96\linewidth]{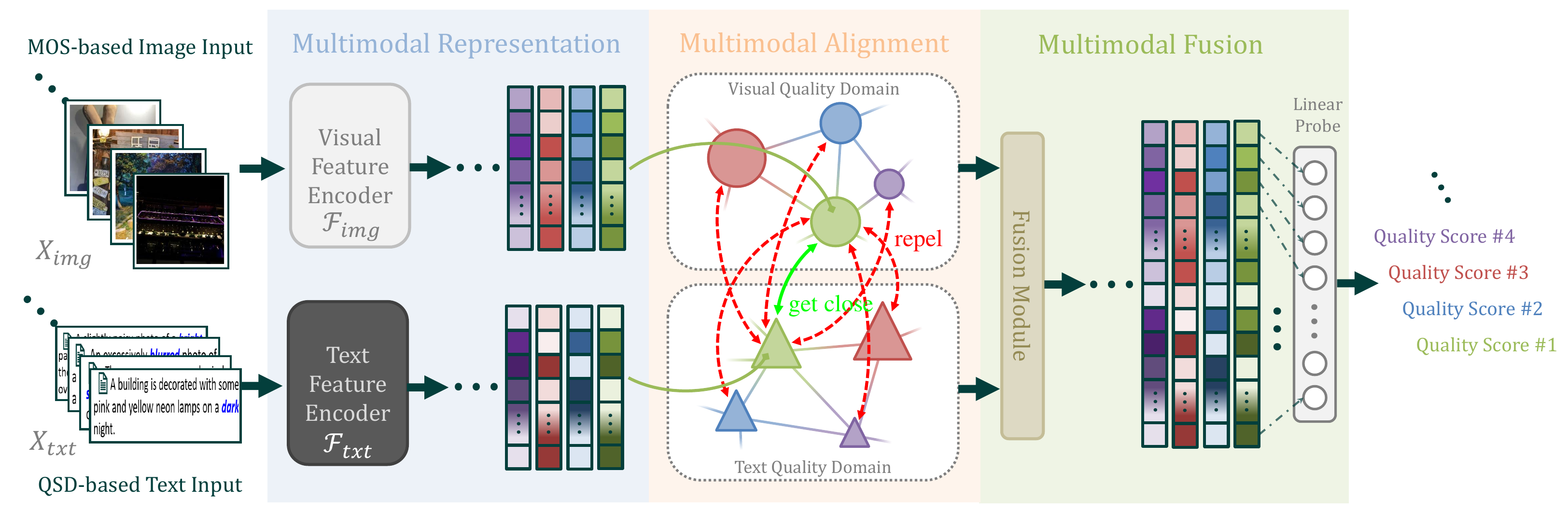}\\
	\caption{\textbf{Overall learning framework of the proposed BMQA}. It consists of three key modules, including multimodal feature representation, latent feature alignment, and fusion prediction.}
	\label{fig:bmqa_framework}
\end{figure}

Based on the proposed MLIQ database, we design a unique deep-learned BMQA  as shown in Fig.~\ref{fig:bmqa_framework}. We address the main challenges of multimodality learning in the BIQA task, including feature representation, alignment, and fusion.
Finally, we will describe our learning mechanism.

\subsection{Multimodal Quality Representation}\label{subSec:multimodal_representation}
In this section, multimodal quality representation refers to extracting and integrating effective features that take the advantage of the supplementary quality description clues. 
Due to the fact that heterologous data has significantly different characteristics, this indicates a large difference between image and text spaces \cite{song2019harmonized}. Therefore, we design two different feature quality representations for them.

\subsubsection{Text Quality Feature Representation}
In this subsection, we explore the quality cognition in terms of the text feature. The text quality representation $\mathbf{F}_{txt}$ can be defined by 
\begin{equation}\label{eqn:text_feature_encoder}
\small
\vspace{-3pt}
\mathbf{F}_{txt}=\mathcal{F}_{txt}\left(X_{txt} {;~} \boldsymbol{\theta}_{txt} \right), 
\end{equation} 
where $\mathcal{F}_{txt}$ represents a text feature extractor, $X_{txt}$ denotes an input quality description sentence, and $\boldsymbol{\theta}_{txt}$ represents the corresponding weights.  $X_{txt}$=$<$$x_{1},\cdots,x_{N}$$>$ denotes the spoken words with a length of $N$.

Next, we extract the latent text feature from $X_{txt}$. Given a word $x_{k}\in<x_{1},\cdots,x_{N}>$, we adopt $\mathcal{F}_{text}$ to extract the semantic quality feature for each word. 
In BMQA, $\mathcal{F}_{txt}$ can be formulated as follows:
\begin{equation}\label{eqn:text_basic_function} 
\small
\vspace{-3pt}
\begin{aligned}
		\mathcal{F}_{txt}\left(x_{k}\right)=&\sigma\left[\mathcal{F}^{1}_{lp}\left(x_{n},<x_{1},\cdots,x_{n},0,\cdots>{;~}\bm{\theta}^{1}_{lp}\right)\right] \\
		&\quad\oplus\mathcal{F}^{2}_{lp}\left(x_{n},<x_{1},\cdots,x_{n},0,\cdots>{;~}\bm{\theta}^{2}_{lp}\right),
\end{aligned}
\end{equation}
where $\mathcal{F}^{1}_{lp}$ and $\mathcal{F}^{2}_{lp}$ denote two linear projections (\textit{e.g.}, fully connection or attention projection), $\bm{\theta}^{1}_{lp}$ and $\bm{\theta}^{2}_{lp}$ denote the corresponding weights, $\sigma$ denotes a nonlinear activation function (\textit{e.g.}, rectified linear unit or Gaussian error linear unit), and $\oplus$ denotes a fusion operation (\textit{e.g.}, addition or concatenation). 
Currently, there is still no consensus on the efficient semantic processing \cite{stiennon2020learning}. The main reason is that different methods build different word associations (\textit{i.e.}, different settings of $\mathcal{F}_{lp}^1$,  $\mathcal{F}_{lp}^2$, $\sigma$ and $\oplus$), and they show different advantages in natural language processing (NLP).

Finally, the association of all words in a dictation will contribute to the feature representation of $\mathbf{F}_{txt}$.
In BMQA, it is generated by a linear combination of all word features:
\begin{equation}\label{eqn:output_text_feature} 
\small
\vspace{-3pt}
\begin{aligned}
\mathbf{F}_{txt}=\mathcal{F}_{lc}\left[\mathcal{F}_{txt}\left(x_{1}\right),\cdots,\mathcal{F}_{txt}\left(x_{N}\right){;~}\bm{\theta}_{lc}\right],
\end{aligned}
\end{equation}
where $\mathcal{F}_{lc}$ represents a fully-connected layer and $\bm{\theta}_{lc}$ denotes the weights of $\mathcal{F}_{lc}$. In the experiments, we select 3 representative networks as the backbone of $\mathcal{F}_{txt}$, including Bag-of-Word (\textit{denoted as} BoW), recurrent neural network (\textit{denoted as} RNN), and Transformer (\textit{denoted as} TransF).

\subsubsection{Image Quality Feature Representation}
Recent work has demonstrated the feasibility and superiority of deep  neural network (DNN) in many vision tasks. Inspired by this, we obtain the image quality representation $\mathbf{F}_{img}$ by a deep-learned mapping:
\begin{equation}\label{eqn:output_visual_feature}
\small
\vspace{-3pt}
\mathbf{F}_{img}=\mathcal{F}_{img}\left(X_{img} {;~} \boldsymbol{\theta}_{img} \right), 
\end{equation} 
where $\mathcal{F}_{img}$ denotes an image feature encoder, $X_{img}$ denotes an input image, and $\boldsymbol{\theta}_{img}$ represents the corresponding weights.

Existing deep-learned image quality representation \cite{fang2022perceptual} mainly relies on the extraction and integration of features in the spatial domain. $\mathcal{F}_{img}$ is usually designed as a stack of multiple layers of DNN blocks. Specifically, let $\mathcal{F}^{l}_{img}$ denote a basic DNN block in the $l$-th layer of $\mathcal{F}_{img}$. $\mathcal{F}^{l}_{img}$ usually consists of sequential linear computations and nonlinear activation functions in BMQA, which can be formulated as:
\begin{equation}\label{eqn:output_image_feature} 
\small
\vspace{-3pt}
\begin{aligned}
		\mathcal{F}^{l}_{img}\left({X}^{l}_{img}\right)=\sigma_{l}\left[\mathcal{F}^{1}_{lp}\left({X}^{l}_{img}{;~}\bm{\theta}^{1}_{lp}\right)\right]\oplus\mathcal{F}^{2}_{lp}\left({X}^{l}_{img}{;~}\bm{\theta}^{2}_{lp}\right),
\end{aligned}
\end{equation}
where $\mathcal{F}^{1}_{lp}$ and $\mathcal{F}^{2}_{lp}$ denote two linear projections (\textit{e.g.}, fully connection, attention projection, or  convolution), $\bm{\theta}^{1}_{lp}$ and $\bm{\theta}^{2}_{lp}$ denote the corresponding weights, $\sigma$ denotes a nonlinear activation function (\textit{e.g.}, rectified linear unit or gaussian error linear units), and $\oplus$ denotes a fusion operation (\textit{e.g.}, addition or concatenation).

Recent studies demonstrate that different backbones of deep models result in different learning capabilities. It indicates that network architectures may have important impacts on the extraction of $\mathbf{F}_{img}$. Therefore, we select 5 representative networks as the backbone of $\mathcal{F}_{img}$, including VGG, ResNet (\textit{denoted as} RN), EfficientNet (\textit{denoted as} EN), Vision-in-Transformer (\textit{denoted as} ViT), and ConvNeXT (\textit{denoted as} CNXT).

\subsection{Image and Text Quality Feature Alignment}
\label{subsec:multimodal_qulaity_alighment}
In BMQA, multimodal quality alignment refers to finding the corresponding quality representation relationship between two modalities. The image and text are obtained for the same image, and the quality description can be highly consistent  \cite{chen2022multimodal}. For example, the text keywords can directly indicate the image regions that the visual attention focuses on \cite{liu2022focus}, thus improving the learning performance of an image feature encoder.

Existing methods build the multimodal alignment by designing constraints across different modalities (called cross-modal constraints \cite{baltruvsaitis2018multimodal}). In the construction of training objective, it is necessary to define a special metric to measure the  difference between two modalities. If two modalities come from different perspectives of a single sensor, the metric is defined as an absolute value error, such as mean absolute error (MAE) and mean square error (MSE). If two modalities come from different sensors, the metric is defined as a relative value error, such as cosine similarity \cite{radford2021learning}.

The image and text are heterogeneous modalities. 
Therefore, we adopt the cosine similarity to measure the relative difference, and design an attentive pooling for multimodal quality alignment. Specifically, we find the shared quality information by learning the attentive distribution, and the aligned visual feature $\widehat{\mathbf{F}}_{img}$ is formulated by:
\begin{equation}\label{eqn:image_attentive_embedding}
\small
\left\{ {\begin{array}{*{20}l}
   {\widehat{{\mathbf{F}}}_{img}=\mathcal{N}_{ln}\left(\mathcal{F}_{\mathrm{fw}}\left(\ddot{{\mathbf{F}}}_{img}{;~}\boldsymbol{\theta}_{\mathrm{fw}}\right)+\ddot{{\mathbf{F}}}_{img}\right);}  \\
   {\ddot{{\mathbf{F}}}_{img}=\mathcal{N}_{ln}\left(\mathcal{F}_{\mathrm{ms}}\left({{\mathbf{F}}}_{img}{;~}\boldsymbol{\theta}_{\mathrm{k}}, \boldsymbol{\theta}_{\mathrm{q}},\boldsymbol{\theta}_{\mathrm{v}},\sigma\right)+{{\mathbf{F}}}_{img}\right)}  \\
\end{array}} \right.,
\vspace{-3pt}
\end{equation}
where $\mathcal{F}_{\mathrm{fw}}$ denotes a feed-forward layer with the parameter weights $\boldsymbol{\theta}_{\mathrm{fw}}$, and $\mathcal{N}_{ln}$ denotes the layer normalization. $\mathcal{F}_{ms}$ denotes a $32$-head self-attention pooling module. 
$\boldsymbol{\theta}_{\mathrm{k}}$, $\boldsymbol{\theta}_{\mathrm{q}}$, and $\boldsymbol{\theta}_{\mathrm{v}}$ denote the parameter weights of key, query and value embedding in $\mathcal{F}_{ms}$, and we set the projection dimension size as $2048$.
{\small$\sigma\left(\cdot\right)=\texttt{Softmax}(\cdot/\sqrt{d})$} denotes a scaled \texttt{softmax} used in $\mathcal{F}_{\mathrm{ms}}$.
Eq. \eqref{eqn:image_attentive_embedding} indicates that the feature elements related to shared quality information will obtain higher attention values than others. Therefore, the aligned quality information will be a weighted combination of all feature elements.

The aligned text feature $\widehat{{\mathbf{F}}}_{txt}$ can be formulated as: 
\begin{equation}\label{eqn:text_attentive_embedding}
\vspace{-3pt}
	\small
	\left\{ {\begin{array}{*{20}l}
		\widehat{{\mathbf{F}}}_{txt}=\mathcal{N}_{ln}\left(\mathcal{F}_{\mathrm{fw}}\left(\ddot{{\mathbf{F}}}_{txt}{;~}\boldsymbol{\theta}^{_{'}}_{\mathrm{fw}}\right)+\ddot{{\mathbf{F}}}_{txt}\right); \\
		\ddot{{\mathbf{F}}}_{txt} =\mathcal{N}_{ln}\left(\mathcal{F}_{\mathrm{ms}}\left({{\mathbf{F}}}_{txt}{;~} \boldsymbol{\theta}^{_{'}}_{\mathrm{k}},\boldsymbol{\theta}^{_{'}}_{\mathrm{q}},\boldsymbol{\theta}^{_{'}}_{\mathrm{v}},\sigma\right)+{{\mathbf{F}}}_{txt}\right)
	\end{array}} \right.,
\end{equation}
where $\boldsymbol{\theta}^{_{'}}_{\mathrm{k}}$, $\boldsymbol{\theta}^{_{'}}_{\mathrm{q}}$, and $\boldsymbol{\theta}^{_{'}}_{\mathrm{v}}$ denote the embedding parameter weights. $\boldsymbol{\theta}^{_{'}}_{\mathrm{fw}}$ denotes the feed-forward parameter weights for the    text feature alignment.

Finally, the relative difference between two aligned $1024$-wide $\widehat{\mathbf{F}}_{img}$ and $\widehat{\mathbf{F}}_{txt}$ is measured by cosine similarity, which can be formulated by
\begin{equation}\label{eqn:text_visual_distance}
\vspace{-3pt}
\small
\mathcal{D}_{cos}\left(\widehat{\mathbf{F}}_{img},\widehat{\mathbf{F}}_{txt}\right)=\frac{\widehat{\mathbf{F}}_{img}\odot \widehat{\mathbf{F}}_{txt}}{\left\|\widehat{\mathbf{F}}_{img}\right\|\left\|\widehat{\mathbf{F}}_{txt}\right\|},
\end{equation}
where $\odot$ represents an inner product operation and $\left\|\cdot\right\|$ represents the Euclidean distance. 
{\small$\mathcal{D}_{cos}\left(\widehat{\mathbf{F}}_{img},\widehat{\mathbf{F}}_{txt}\right)$} will be used as the metric of multimodal self-supervision learning in Sec. \ref{subsec:multimodal_learning_mechanism}.

\subsection{Image and Text Quality Feature Fusion}
\label{subsec:multimodal_qulaity_fusion}
In BMQA, the main benefit of using two modalities is that the image quality can be described from different perspectives \cite{d2015review}. For example, quality scores can be obtained directly from subjective tests, which can be reflected by the text information. Another representative case is that verbal descriptions of highly similar scenes are often unavoidably similar, while spatial details from images will help fine-grained scoring decisions. These two scenarios show that the fusion of image and text quality representations helps to predict more accurate quality scores. Therefore, multimodal quality fusion \cite{baltruvsaitis2018multimodal} is adopted to integrate two modalities to predict the quality score in our BMQA.

To preserve quality information as much as possible, we integrate two heterogenous modality features via a concatenation operation. Next, we employ a $2048$-wide linear probe \cite{radford2021learning}, $\mathcal{F}_{fuse}$, to fuse and forecast a final quality score $s_{pred}$, which can be formulated as:
\begin{equation}\label{eqn:fusion}
\vspace{-3pt}
\small
	s_{pred}=\mathcal{F}_{fuse}\left[\texttt{concat}\left(\widehat{\mathbf{F}}_{img}, \widehat{\mathbf{F}}_{txt}\right){;~}\boldsymbol{\theta}_{fuse}\right],
\end{equation}
where $\texttt{concat}\left(\cdot\right)$ denotes the concatenation operation, and $\boldsymbol{\theta}_{fuse}$ denotes the parameter weights of $\mathcal{F}_{fuse}$.

\subsection{Deep Multimodal Learning Mechanism}
BMQA learns to predict image quality scores by exploiting the supplementary quality descriptions from cross-modal features. In this section, we describe our multimodal learning mechanism, including multimodal self-supervision and supervision.
\begin{algorithm}[!t]
	\small
	\caption{Proposed BMQA scheme on the image-text case, BMQA$^{image\raisebox{-0.35mm}{-}text}$.}\label{algm:bmqa}
	\hspace*{0.02in} {\bf Input:} \\
	\hspace*{0.25in}  Image sample $X_{img}$ and quality description text $X_{txt}$. \\
	\hspace*{0.02in} {\bf Output:} \\
	\hspace*{0.25in}   Quality score $s_{pred}$.
	\begin{algorithmic}[1]
		\State Extract the text feature representation $\mathbf{F}_{txt}$ in Eqs. \eqref{eqn:text_basic_function}-\eqref{eqn:output_text_feature};
		\State Extract the image feature $\mathbf{F}_{img}$ in Eqs. \eqref{eqn:output_visual_feature}-\eqref{eqn:output_image_feature}; 
		\State Obtain the aligned image feature $\widehat{\mathbf{F}}_{img}$ in Eq. \eqref{eqn:image_attentive_embedding} and the aligned text feature in $\widehat{\mathbf{F}}_{txt}$ in Eq. \eqref{eqn:text_attentive_embedding}, respectively;
		\State Fuse $\widehat{\mathbf{F}}_{img}$ and $\widehat{\mathbf{F}}_{txt}$, and predict the objective quality score $s_{pred}$ in Eq. \eqref{eqn:fusion}.
	\end{algorithmic}
\end{algorithm}

\subsubsection{Multimodal Quality Self-Supervision}
\label{subsec:multimodal_learning_mechanism}
Existing methods suggest that image quality status can be latently learned without subjective scores. 
In other words, quality status can be captured and learned from distorted image samples. The multimodal information is expected to enhance deep model learning, since the supplementary information from two modalities will contribute to the quality representation  learning. For example, people prefer `\textit{bright}' images compared with `\textit{dark}' images.

The shared quality information of two modalities from the same instance provides supplementary information.  Therefore, we learn an embedding space, in which feature pairs from the same instance gets close while these from different instances get far away from each other. Specifically, given the $i$-th $\hat{{\mathbf{F}}}_{img(i)}$ in the current training batch $B$, the pair-wise probability $\mathcal{P}\left(\cdot\right)$ is calculated to find the matched $j$-th text feature as:
\begin{equation}\label{eqn:text2image_pairwise_probability}
\vspace{-3pt}
\small
\mathcal{P}_{img}\left(i,j\right)=\frac{\exp\left(\mathcal{D}_{cos}\left(\widehat{{\mathbf{F}}}_{img(i)},\widehat{{\mathbf{F}}}_{txt(j)}\right)/\tau\right)}
		{\sum\nolimits_{k\in{B}}\exp\left(\mathcal{D}_{cos}\left(\widehat{{\mathbf{F}}}_{img(i)},\widehat{{\mathbf{F}}}_{txt(k)}\right)/\tau\right)},
\end{equation}
where $\tau$ denotes the temperature parameter that controls the degree of distribution concentration.

At the same time, the pair-wise probability to find the matched $i$-th image feature for the $j$-th text feature $\hat{{\mathbf{F}}}_{txt(j)}$ will be:
\begin{equation}\label{eqn:image2text_pairwise_probability}
\vspace{-3pt}
\small
\mathcal{P}_{txt}\left(j,i\right)=\frac{\exp\left(\mathcal{D}_{cos}\left(\widehat{{\mathbf{F}}}_{img(i)},\widehat{{\mathbf{F}}}_{txt(j)}\right)/\tau\right)}
		{\sum\nolimits_{k\in{B}}\exp\left(\mathcal{D}_{cos}\left(\widehat{{\mathbf{F}}}_{img(k)},\widehat{{\mathbf{F}}}_{txt(j)}\right)/\tau\right)}.
\end{equation}

Therefore, multimodal quality alignment needs to satisfy that the pair-wise probability between each paired $\widehat{{\mathbf{F}}}_{img}$ and $\widehat{{\mathbf{F}}}_{txt}$ gets the maximum value as: $\mathcal{P}_{img}\left(i,i\right)=\max\nolimits_{k\in{B}}\mathcal{P}_{img}\left(i,k\right)$ and $\mathcal{P}_{txt}\left(j,j\right)=\max\nolimits_{k\in{B}}\mathcal{P}_{txt}\left(j,k\right)$.
The overall learning goal is to maximize the joint probability, which is equivalent to minimizing the negative log-likelihood:
\begin{equation}\label{eqn:self_supervision_loss}
\small
\mathop{\min}\limits_{_{\substack{\boldsymbol{\theta}_\mathrm{img},\;\boldsymbol{\theta}_\mathrm{fw},\boldsymbol{\theta}_\mathrm{k},\boldsymbol{\theta}_\mathrm{q},\boldsymbol{\theta}_\mathrm{v};\\\boldsymbol{\theta}_\mathrm{txt},\;\boldsymbol{\theta}^{_{'}}_\mathrm{fw},\boldsymbol{\theta}^{_{'}}_\mathrm{k},\boldsymbol{\theta}^{_{'}}_\mathrm{q},\boldsymbol{\theta}^{_{'}}_\mathrm{v}}}}
 -\log\left[\sum\limits_{i\in{B}}\mathcal{P}_{img}\left(i,i\right)+\sum\limits_{j\in{B}}\mathcal{P}_{txt}\left(j,j\right)\right].
\end{equation}

\subsubsection{Multimodal Quality Supervision}
To obtain a better fusion model $\mathcal{F}_{fuse}$ in Eq. \eqref{eqn:fusion}, we adopt the mean square error $\left\|\cdot\right\|^2_2$ to measure the difference between the predicted score $s_{pred}$ and the ground-truth subjective score $s_{gt}$. Our goal is to predict a score as close as possible to the subjective MOS, and the optimization problem can be formulated as:
\begin{equation}\label{eqn:supervision_loss}
\small
\mathop{\min}\limits_{\boldsymbol{\theta}_{fuse}}\left\|\mathcal{F}_{fuse}\left[\texttt{concat}\left(\widehat{{\mathbf{F}}}_{img}, \widehat{{\mathbf{F}}}_{txt}\right){;~}\boldsymbol{\theta}_{fuse}\right] -s_{gt}\right\|_2^2.
\end{equation}

\subsection{Overall Algorithm}
Given a pair of image data $X_{img}$ and text data $X_{txt}$, we obtain the image feature $\mathbf{F}_{img}$ by Eq. \eqref{eqn:output_visual_feature} and the text feature $\mathbf{F}_{txt}$ by Eq. \eqref{eqn:output_text_feature}. After that, BMQA learns to output the aligned features $\widehat{\mathbf{F}}_{img}$ and $\widehat{\mathbf{F}}_{txt}$ by Eq. \eqref{eqn:image_attentive_embedding} and Eq.  \eqref{eqn:text_attentive_embedding}, respectively. The learning of the quality feature alignment will be achieved by a  multimodal self-supervision. Finally, BMQA learns to fuse the aligned feature by Eq. \eqref{eqn:fusion}, where the learning of the quality feature fusion will be achieved by a multimodal supervision.

The detailed algorithm of our BMQA is summarized in \textbf{Algorithm} \ref{algm:bmqa}.

\section{Validations and Discussions}
In this section, extensive experiments are conducted on two latest benchmark low-light image databases.  Specifically, we verify the effectiveness of our BMQA on the image-text database MLIQ (\textit{i.e.}, BMQA$^{image\raisebox{-0.35mm}{-}text}$) and the image-only database Dark-4K (\textit{i.e.}, BMQA$^{image\raisebox{-0.35mm}{-}only}$), respectively. Besides, we demonstrate the comparison results with {25 competitive methods}, including 8 hand-crafted BIQAs and 17 deep-learned BIQAs. We provide the detailed descriptions of the experimental validation, analysis, and discussion as follows.

\subsection{Experimental Protocols}
\label{subsec:experimental_protocols}

\subsubsection{Benchmark Database}\label{subSec:benchmark_database}
In the experiments, three databases are used including a pre-training database MS-COCO, a new multimodal image quality database MLIQ, and a new low-light database Dark-4K containing only a single image modality.

\vspace{3pt}
\noindent\textbf{MS-COCO}.
We pre-train our BMQA model on the MS-COCO caption database \cite{vinyals2016show}, which contains $82783$ image instances labeled with captions. The sentence of each caption contains at least 8 words. The partitions of MS-COCO strictly follow the officially specified setting.

\vspace{3pt}
\noindent\textbf{MLIQ}.
We train our BMQA model on the proposed MLIQ dataset, containing $3600$ pairs of low-light image and corresponding text description. The details can be found in Sec. \ref{sec:database_constrution}. In the experiments, MLIQ is randomly divided by \textbf{8:1:1} according to the shooting scene. Specifically, the training set contains {576 scenes and 2880 samples, the validating set contains 72 scenes and 360 samples, and the testing set also contains 72 scenes and 360 samples.}

\vspace{3pt}
\noindent\textbf{Dark-4K}. 
To validate the cross-dataset performance, we further establish a new ultra-high-definition (UHD) low-light database for the cross-dataset validation, namely Dark-4K. The original images of Dark-4K are collected from  \cite{chen2018learning}. 
Dark-4K consists of $424$ raw low-light images, which are captured by two consumer electronics: \textit{Sony $\alpha$7S-II} and \textit{Fujifilm X-T2}. These two cameras have different imaging hardware: \textit{Sony} has a full-frame Bayer sensor, and  \textit{Fujifilm} has an APS-C X-Trans sensor. Dark-4K supports the quality assessment of low-light images produced by different filter arrays. 

\begin{figure}[!t]
\raggedright
\begin{tabular}{p{3.75cm}<{\centering} p{3.75cm}<{\centering}}
\includegraphics[width=1.1\linewidth, height=0.60\linewidth]{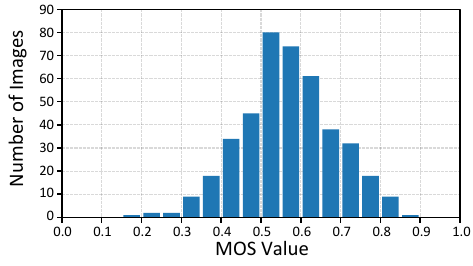} 
&\includegraphics[width=1.1\linewidth, height=0.60\linewidth]{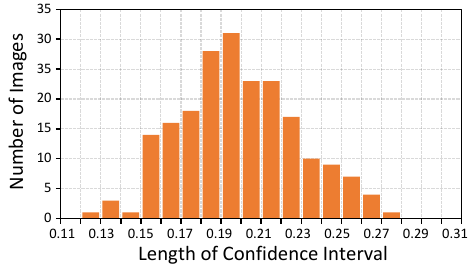} \\  
{~~~~~~~~}(a) {\scriptsize MOS distribution} & {~~~~~~~~}(b) {\scriptsize Confidence interval} \\  	
\end{tabular}    
\vspace{3pt}
\caption{\textbf{A statistical analysis of the proposed unimodal Dark-4K database}: (a) histogram of  MOS values and (b) distribution of confidence interval.}
\label{fig:dark_4k_statistics}
\end{figure}

The subjective experiments on Dark-4K maintain the same settings as described in Sec. \ref{subsubsec:image_modalilty}. The histogram of labeled MOS results and the $95\%$ confidence intervals for the subjective ratings are shown in Fig.~\ref{fig:dark_4k_statistics} (a) and Fig.~\ref{fig:dark_4k_statistics} (b), respectively.

\vspace{-0.1cm}
\subsubsection{Evaluation Metric}
In visual quality assessment, Pearson Linear Correlation (PLCC), Spearman Rank Correlation Coefficient (SRCC), and Root Mean Square Prediction Error (RMSE) are three commonly-used evaluation metrics  \cite{zhang2022continual}. PLCC is used to measure the linear relationship between objective predictions and subjective scores, SRCC reflects the monotonicity of predictions, and RMSE measures the accuracy of predictions. For a  promising method, PLCC and SRCC are close to 1, while RMSE is close to 0.

\vspace{-0.1cm}
\subsubsection{Training Detail}\label{subSec:training_detail}
All experiments have been carried out on a computing server with \textit{Intel(R) Xeon(R) Gold 6226R} CPU\textit{@2.90GHz}, 38GB RAM, and \textit{NVIDIA A100-PCIE} GPU\textit{@40GB}$\times$6.

Next, we report the setting of each training stage, such as 1) pre-training, 2) self-supervised training, and 3) supervised training. Note that \textit{Adam} is used as our optimization solver in \textit{Python Toolbox PyTorch}. 

\vspace{3pt}
\noindent\textbf{{Self-supervised} pre-training} (Stage$_{_{\mathrm{PT}}}$).
We first pre-train the image feature extractor $\mathcal{F}_{img}$ as well as {the text   feature extractor} $\mathcal{F}_{txt}$ on the MS-COCO database. We take $768$ samples as a batch (\textit{i.e.}, 128$\times$6) and pre-train our models for 50 epochs. 
We set the initial learning rate to $1e$-$3$ and decay it by a \textit{cosine} schedule.

\vspace{3pt}
\noindent\textbf{Self-supervised training} (Stage$_{_{\mathrm{SS}}}$).
We train our BMQA model using the pairs of low-light image and the corresponding QSD-based text from the proposed MLIQ database.
At this time, we reduce the batch size to 256 and set the total training epochs to 20, where the initial learning rate is fixed at $4e$-$5$. In the  experiments, we find that neither undertrained (\textit{e.g.}, a loss greater than 1.0) nor overtrained (\textit{e.g.}, a loss less than 0.6) models can achieve better performance. Empirically, we randomly pick the final model with the loss around 0.8.

\vspace{3pt}
\noindent\textbf{Supervised training} (Stage$_{_{\mathrm{ST}}}$).
For VGG, ResNet and EfficientNet, we set the initial learning rate to $8e$-$5$ and reduce the learning rate by a factor of $0.95$ at the \textit{150}-th and \textit{250}-th epochs.
For ViT and ConvNeXT, we set a smaller initial learning rate to $6.4e$-$5$ and also reduce the learning rate by a factor of $0.95$ at the \textit{150}-th and \textit{250}-th epochs.
Finally, we keep the batch size as 16 and train the quality score predictor for 300 epochs.
\begin{table}[!t]
\centering
\footnotesize
\caption{\textbf{Overall performance comparison between 15 $\mathcal{F}_{img}$ and $\mathcal{F}_{txt}$ combinations on the MLIQ database}.}
\label{tab:overall_performance_different_net}
\renewcommand{\arraystretch}{1.15}
\scriptsize
\setlength{\tabcolsep}{4.5mm}{
\begin{tabular}{cc|ccc}
			\toprule[1.5pt]
			$\mathcal{F}_{img}$&$\mathcal{F}_{txt}$&PLCC $\uparrow$ &SRCC $\uparrow$ &RMSE $\downarrow$\\
			\hline
			\multirow{3}{*}{VGG-19}			
			&BoW&0.8538&0.8532&0.0885\\ 
			&RNN&0.8670&0.8636&0.0869\\ 
			&TransF&0.8731&0.8685&0.0847 \\
			\hline
			\multirow{3}{*}{RN-50}
			&BoW&0.8743&0.8750&0.0845 \\ 
			&RNN&0.8859&0.8851&0.0824\\ 
			&TransF&0.8989&0.8877&0.0830\\
			\hline
			\multirow{3}{*}{EN-B4}
			&BoW&0.8760&0.8719&0.0834 \\ 
			&RNN&0.8898&0.8886&0.0810 \\ 
			&TransF&0.8987&0.8922&0.0825 \\
			\hline
			\multirow{3}{*}{ViT-B32}
			&BoW&0.8802&0.8774&0.0827 \\ 
			&RNN&0.8879&0.8871&0.0822 \\ 
			&TransF&0.9089&0.9040&0.0816 \\
			\hline
			\multirow{3}{*}{CNXT-B}
			&BoW&0.8822&0.8844&0.0832 \\ 
			&RNN&0.8970&0.8984&0.0817 \\ 
			&TransF&0.9121&0.9065&0.0802 \\
			\bottomrule[1.5pt]
\end{tabular}}
\end{table}

\subsection{Feature Representation Validation}\label{subSec:framework_generalization_validation} 
To verify the effectiveness of two heterogenous modality feature extractors, we have conducted  the experiments on several representative $\mathcal{F}_{txt}$ and $\mathcal{F}_{img}$ models. Specifically, we take VGG-19, ResNet-50 (\textit{denoted as} RN-50), EfficientNet (\textit{denoted as} EN-B4), Vision-in-Transformer (\textit{denoted as} ViT-B32), and ConvNeXT (\textit{denoted as} CNXT-B) as the image feature encoders,  respectively. In addition,  we take Bag-of-Word (\textit{denoted as} BoW), recurrent neural network (\textit{denoted as} RNN), and Transformer (\textit{denoted as} TransF) as the text feature encoders, respectively.

Considering that the prediction accuracy of BMQA is mainly related to the feature representation, we first explore the impact of different network backbones of $\mathcal{F}_{img}$, including VGG19, RN-50, EN-B4, ViT-B32, and CNXT-B. Meanwhile, we employ BoW, RNN, and TransF as $\mathcal{F}_{txt}$. Table \ref{tab:overall_performance_different_net} provides the overall comparison results of $15$ $\mathcal{F}_{txt}$ and $\mathcal{F}_{img}$ combinations. As seen, all 15 BMQA variants achieve promising performance, which verify the excellent robustness capability of our multimodal paradigm. Besides, the related combinations achieve higher prediction accuracy with the use of more powerful feature encoders.

\begin{figure}[!t]
\raggedright
\begin{tabular}{p{3.75cm}<{\centering} p{3.75cm}<{\centering}}
\includegraphics[width=1.1\linewidth, height=1.20\linewidth]{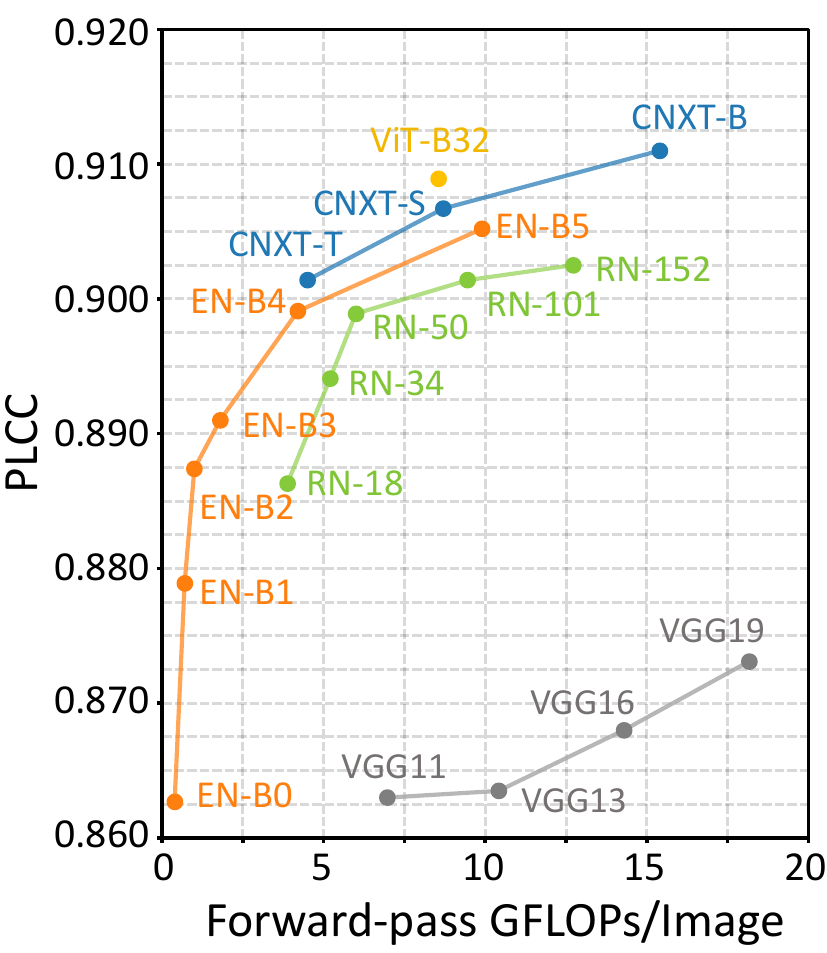} 
&\includegraphics[width=1.1\linewidth, height=1.20\linewidth]{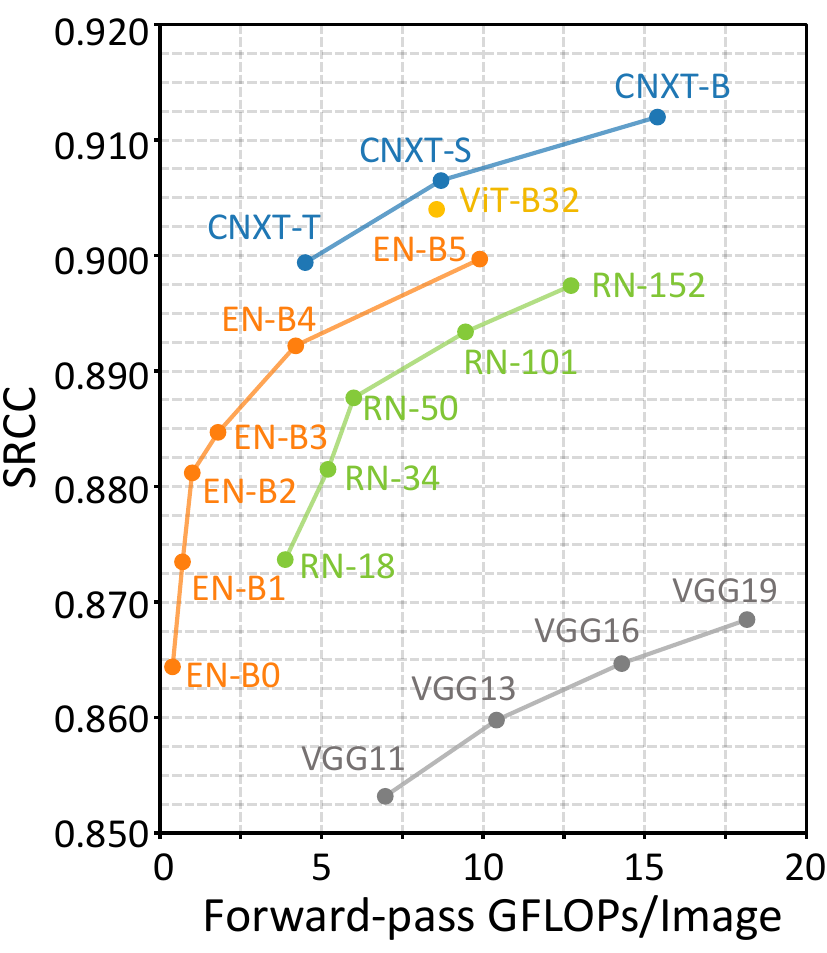} \\  
{~~~~~~~~~~}(a) {\scriptsize PLCC-GFLOPs} & {~~~~~~~~~~}(b) {\scriptsize SRCC-GFLOPs} \\  	
\end{tabular}    
\vspace{2pt}
\caption{\textbf{Performance comparison of different image feature extractors} $\mathcal{F}_{img}$: (a) PLCC-GFLOPs curves of different network variants and (b) SRCC-GFLOPs curves of different network variants.}
\label{fig:curves_framework_plccsrcc}
\end{figure}

In addition, considering that image is the main modality in the BIQA task, we further explore the impact of different network variants of $\mathcal{F}_{img}$. Fig.~\ref{fig:curves_framework_plccsrcc} provides the PLCC and SRCC curves of five $\mathcal{F}_{img}$ models. 
As seen, we can draw some interesting conclusions: 
1) Different variants of the same network architecture have a little effect on the final forecasting results. For example, the replacement of the ResNet network (from the heaviest RN-152 to the lightest RN-18) only results in the PLCC decreasing from {$0.8863$ to $0.9025$} and the SRCC decreasing from {$0.8737$ to $0.8974$};  
2) The lightweight models have also achieved competitive  performance. For example, EN-B0 achieves a PLCC score of {$0.8627$ and a SRCC score of $0.8532$} with only 0.39G FLOPs. 
This suggests that our BMQA framework can maintain excellent performance for lightweight models.

\subsection{Ablation Study}\label{subSec:ablation_study}
As mentioned above, BMQA is based on two modalities and is constructed through three training phases, including pre-training, multimodal self-supervision, and multimodal supervision. In view of this, the ablation experiments are conducted to verify the effectiveness of two different modalities for different learning stages. The detailed results are provided in Table~\ref{tab:ablation_study}.

\vspace{3pt}
\noindent\textbf{Multimodal Self-supervision.}
To demonstrate the roles of two modalities, we first verify the performance of a single modality in Method (a) and Method (b), where ``-'' means the related modality or training is not used in Table~\ref{tab:ablation_study}. In other words, BMQA independently adopts RN-50 as the image encoder $\mathcal{F}_{img}$ or TransF as the text encoder $\mathcal{F}_{txt}$.  
As seen, the result of Method (a) is better than that of Method (b), which confirms that the image data is the main modality in the visual quality assessment task.

To verify the effect of Stage$_{_{\mathrm{SS}}}$, we also design Method (c) with the self-supervised learning based on Method (a), which is carried out for the feature alignment between TransF ($\mathcal{F}_{txt}$) and RN-50 ($\mathcal{F}_{img}$)  as described in Sec.~\ref{subsec:multimodal_learning_mechanism}. As seen, Method (c) obtains a better result than Method (a), and it shows that the introduction of multimodal self-supervision helps the deep model training. The same phenomenon can be observed between Methods (d) and (b).

\vspace{3pt}
\noindent\textbf{Multimodal Supervision}. 
To verify the effect of Stage$_{_{\mathrm{ST}}}$, we also conduct the experiments with Method (e), which employs  RN-50 as $\mathcal{F}_{img}$ and TransF as $\mathcal{F}_{txt}$ to perform an end-to-end multimodal quality supervision. 
As seen, Method (e) achieves a better result than Method (c) and Method (d), which indicates that the introduction of multimodal supervised learning improves the performance of BMQA.

\vspace{3pt}
\noindent\textbf{Pre-train Strategy.}
To show the feature alignment performance between $\mathcal{F}_{txt}$ and $\mathcal{F}_{img}$, we go a step further by pre-training Method (e) on the MS-COCO database. 
Specifically, we have performed three different pre-training strategies: 1)  Method (f) verifies the pre-training effect on the classification task  (\textit{denoted as} CL), where RN-50 is pre-trained to distinguish categories of $\mathcal{F}_{img}$ and TransF is pre-trained to distinguish categories of $\mathcal{F}_{txt}$. 2) Method (g) verifies the pre-training effect on the restoration task  (\textit{denoted as} RE), where RN-50 is pre-trained to restore the missing image content in $\mathcal{F}_{img}$ and TransF is pre-trained to fill up the missing word information in $\mathcal{F}_{txt}$. 3) Method (h) verifies the pre-training effect on the feature alignment between $\mathcal{F}_{img}$ and $\mathcal{F}_{txt}$, where RN-50 and TransF are pre-trained to align the image and text features. 
Experiments show that Method (h) obtains a better result than Methods (f) and (g), which indicates that introducing  multimodal quality alignment in the pre-training improves the BIQA performance.
\begin{table}[!t]
\centering
\scriptsize
\caption{\textbf{Ablation study of the contribution of each training  stage}:  self-supervised pre-training Stage$_{_{\mathrm{PT}}}$, self-supervised training Stage$_{_{\mathrm{SS}}}$, and supervised training Stage$_{_{\mathrm{ST}}}$.}
\label{tab:ablation_study}
	\renewcommand{\arraystretch}{1.15}
	\setlength{\tabcolsep}{0.95mm}{
		\begin{tabular}{c|c|cc|cc|ccc}
			\toprule[1.4pt]
			\multirow{2}{*}{Method}&\multirow{2}{*}{Stage$_{_{\mathrm{PT}}}$}&\multicolumn{2}{c|}{Stage$_{_{\mathrm{SS}}}$}& \multicolumn{2}{c|}{Stage$_{_{\mathrm{ST}}}$}&\multirow{2}{*}{PLCC$\uparrow$}&\multirow{2}{*}{SRCC$\uparrow$}&\multirow{2}{*}{RMSE$\downarrow$}\\
			\cline{3-6}
			&&$\mathcal{F}_{img}$&$\mathcal{F}_{txt}$&$\mathcal{F}_{img}$&$\mathcal{F}_{txt}$&&&\\
			\hline
			(a)&-&-&-&RN-50&-&0.7964&0.7934&0.1082\\
			(b)&-&-&-&-&TransF&0.7504&0.7466&0.1130\\
			(c)&-&RN-50&TransF&RN-50&-&0.8567&0.8556&0.0902\\
			(d)&-&RN-50&TransF&-&TransF&0.8209&0.8183&0.0932\\
			(e)&-&RN-50&TransF&RN-50&TransF&0.8614&0.8610&0.0896\\
			(f)&CL&RN-50&TransF&RN-50&TransF&0.8707&0.8709&0.0868\\
			(g)&RE&RN-50&TransF&RN-50&TransF&0.8753&0.8721&0.0871\\
			(h)&FM&RN-50&TransF&RN-50&TransF&0.8989&0.8877&0.0830\\
			\bottomrule[1.5pt]
		\end{tabular}}
\end{table}

\begin{table*}[!t]
\caption{\textbf{Performance comparison of the proposed BMQA method and $25$ state-of-the-art BIQAs on the MLIQ database}. The best results of the hand-crafted and deep-learned BIQAs are highlighted in \textbf{bold} for different devices, and the best results of our BMQA are highlighted in \underline{\textbf{underline}}.}
\label{tab:performance_comparison}
\footnotesize
\centering
\renewcommand{\arraystretch}{1.15}
\setlength{\tabcolsep}{0.4mm}{	
\newsavebox{\tablebox}
\begin{lrbox}{\tablebox}
\begin{tabular}{rc|ccc|ccc|ccc|ccc|ccc|ccc}
			\toprule[1.5pt]
			\multicolumn{2}{c|}{Method Type} & \multicolumn{3}{c|}{Device-I (\textit{Nikon D5300})} & \multicolumn{3}{c|}{Device-II (\textit{iPhone 8plus})} & \multicolumn{3}{c|}{Device-III (\textit{iPad mini2})} &\multicolumn{3}{c|}{Device-IV (\textit{Canon EOS})}&\multicolumn{3}{c|}{Device-V (\textit{Huawei Mate})} & \multicolumn{3}{c}{Entire Database}\\
			\Xhline{1.0pt}
			\rowcolor[gray]{.95} Hand-crafted BIQAs&Type &PLCC$\uparrow$&SRCC$\uparrow$&RMSE$\downarrow$ &PLCC$\uparrow$&SRCC$\uparrow$&RMSE$\downarrow$ &PLCC$\uparrow$&SRCC$\uparrow$&RMSE$\downarrow$ &PLCC$\uparrow$&SRCC$\uparrow$&RMSE$\downarrow$ &PLCC$\uparrow$&SRCC$\uparrow$&RMSE$\downarrow$ &PLCC$\uparrow$&SRCC$\uparrow$&RMSE$\downarrow$\\
           \textit{Mittal2012TIP} \cite{mittal2012no}&GP &0.7833&0.7657&0.1073 &0.7865&0.7695&0.0906 &0.8005&0.7706&0.0802 &0.7273&0.7136&0.1247 &0.7709&0.7500&0.1100 &0.7797&0.7716&0.1095 \\
			
			\textit{Liu2014SPIC} \cite{liu2014no}&GP &0.7848&0.7666&0.1072 &0.7894&0.7710&0.0904 &0.7777&0.7593&0.0904 &0.7301&0.7178&0.1265 &0.7678&0.7454&0.1109 &0.7779&0.7689&0.1101 \\
			
			\textit{Zhang2015TIP} \cite{zhang2015feature}&GP &0.6537&0.6704&0.1303 &0.6888&0.6747&0.1069 &0.7883&0.8328&0.0837 &0.5836&0.5832&0.1496 &0.6736&0.6560&0.1286 &0.6186&0.6346&0.1379 \\
			
			\textit{Li2016SPL} \cite{li2016no}&MD &0.8267&0.8242&0.0961 &0.7626&0.7335&0.0936 &0.7935&0.7651&0.0821 &0.8442&0.8338&0.0986 &0.8767&0.8676&0.0824 &0.8317&0.8273&0.0972\\
			
			\textit{Gu2017TCB} \cite{gu2017no}&CD &0.8486&0.8446&0.0911 &0.7957&0.7918&0.0893 &0.7845&0.7878&0.0844 &0.7721&0.7764&0.1171 &0.7726&0.7658&0.1105 &0.7726&0.7658&0.1105  \\
			
			\textit{Xiang2020TMM} \cite{xiang2020blind}&LL &0.8265&0.8221&0.1003 &\textbf{0.8642}&\textbf{0.8530}&0.0885 &0.8105&0.8004&0.0871 &0.8323&0.8146&0.0730 &\textbf{0.8789}&\textbf{0.8642}&\textbf{0.0881} &0.8063&0.7950&0.1011\\
			
			\textit{Wang2021ICME} \cite{wang2021blind}&LL &0.8619&0.8539&0.0868 &0.8304&0.8179&0.0798 &0.8002&0.7759&0.0805 &0.8584&0.8538&0.0942 &0.8386&0.8234&0.0954 &0.8365&0.8350&0.0969 \\
			
			\textit{Wang2022TII} \cite{wang2022low}&LL &\textbf{0.8647}&\textbf{0.8571}&\textbf{0.0845} &0.8401&0.8373&\textbf{0.0815} &\textbf{0.8336}&\textbf{0.8459}&\textbf{0.0713} &\textbf{0.9085}&\textbf{0.8892}&\textbf{0.0789} &0.8577&0.8484&0.0936 &\textbf{0.8578}&\textbf{0.8488}&\textbf{0.0904} \\
			
			\Xhline{1.0pt}
			\rowcolor[gray]{.95} Deep-learned BIQAs&Type &PLCC$\uparrow$&SRCC$\uparrow$&RMSE$\downarrow$ &PLCC$\uparrow$&SRCC$\uparrow$&RMSE$\downarrow$ &PLCC$\uparrow$&SRCC$\uparrow$&RMSE$\downarrow$ &PLCC$\uparrow$&SRCC$\uparrow$&RMSE$\downarrow$&PLCC$\uparrow$&SRCC$\uparrow$&RMSE$\downarrow$&PLCC$\uparrow$&SRCC$\uparrow$&RMSE$\downarrow$\\
			
			\textit{Kang2014CVPR} \cite{kang2014convolutional}&AS &0.8103&0.8197&0.1580 &0.7954&0.7987&0.1537 &0.6822&0.6992&0.1792 &0.8301&0.8629&0.1681 &0.8783&0.8874&0.1575 &0.8095&0.8167&0.1604\\
			
			\textit{Bosse2018TIP} \cite{bosse2017deep}&AS &0.9077&0.9154&0.0918 &0.7395&0.7198&0.0979 &0.6824&0.6305&0.1281 &0.8695&0.8596&0.1198 &0.8516&0.8615&0.1164 &0.8182&0.8198&0.1070\\
			
			\textit{Ke2021ICCV} \cite{ke2021musiq}&AS &0.8614&0.8464&0.0884 &0.8343&0.8254&\textbf{0.0822} &0.8297&0.8208&0.0738 &0.8956&0.8835&0.0800 &0.8362&0.8312&\textbf{0.0944} &0.8504&0.8487&0.0923\\
			
			\textit{Ying2020CVPR} \cite{ying2020patches}&GS &0.8458&0.8427&0.0926 &\textbf{0.8561}&\textbf{0.8468}&0.0893 &0.8117&0.7953&0.0854 &0.8628&0.8138&\textbf{0.0700} &0.8842&0.8616&0.0846 &0.8450&0.8455&0.0921\\
			
			\textit{Kim2019TNNLS} \cite{kim2019deep}&MT &0.9096&0.9166&0.0863 &0.7424&0.7213&0.0963 &0.6828&0.6305&0.1333 &0.8713&0.8609&0.1213 &0.8515&0.8633&0.1180 &0.8217&0.8234&0.1063\\
			
			\textit{Yan2019TMM} \cite{yan2019naturalness}&MT &0.8244&0.8276&0.1631 &0.8030&0.8105&0.1575 &0.7156&0.6992&0.1755 &0.8690&0.8775&0.1700 &0.8398&0.8535&0.1624 &0.8256&0.8370&0.1639\\
			
			\textit{Zhang2020TCSVT} \cite{zhang2020blind}&MT &0.8720&0.8845&0.1075 &0.7118&0.7141&0.1035 &0.6681&0.5929&0.1250 &0.8423&0.8340&0.1163 &0.7631&0.7558&0.1213 &0.8241&0.8261&0.1035\\
			
			\textit{Wu2020TIP} \cite{wu2020end}&MO &0.8713&0.8786&0.1249 &0.7132&0.7188&0.1163 &\textbf{0.8516}&\textbf{0.8277}&0.0721 &0.9265&\textbf{0.9356}&0.0976 &0.8537&0.8498&0.0989 &0.8299&0.8371&0.1099\\
			
			\textit{Li2020ACMMM} \cite{li2020norm}&MO &0.8561&0.8514&0.0877 &0.7981&0.7894&0.0882 &0.8074&0.8275&\textbf{0.0706} &0.8866&0.8740&0.0869 &0.8174&0.8028&0.0995 &0.8236&0.8223&0.0988\\
			
			\textit{Su2020CVPR} \cite{su2020blindly}&MO &0.8645&0.8581&0.0843 &0.7933&0.7898&0.0791 &0.8171&0.8226&0.0664 &\textbf{0.9303}&0.9250&0.0677 &0.8401&0.8427&0.0893 &0.8520&0.8510&0.0882\\
			
			\textit{Zhu2020CVPR} \cite{zhu2020metaiqa}&MO &0.9248&0.9311&0.0721 &0.7709&0.7626&0.0922 &0.6819&0.6283&0.1211 &0.8890&0.8814&0.1096 &0.8513&0.8603&0.1031 &0.8519&0.8547&0.0948\\
			
			\textit{Ma2021ACMMM} \cite{ma2021remember}&MC &0.8594&0.8623&0.0871 &0.7971&0.7927&0.0877 &0.8049&0.8017&0.0771 &0.8631&0.8592&0.0903 &0.8092&0.7997&0.0987 &0.8271&0.8275&0.0982\\	
			
			\textit{Zhang2021TIP} \cite{zhang2021uncertainty}&MC &0.8848&0.8838&0.0880 &0.8487&0.8466&0.0832 &0.8479&0.8051&0.0708 &0.8476&0.8482&0.1052 &0.8147&0.8135&0.1008 &0.8485&0.8605&0.0929\\
			
			\textit{Zhang2022TPAMI} \cite{zhang2022continual}&MC &\textbf{0.9277}&\textbf{0.9320}&\textbf{0.0713} &0.7808&0.7712&0.0911 &0.6811&0.6403&0.1117 &0.8938&0.8870&0.1037 &\textbf{0.8908}&\textbf{0.8977}&0.0975 &\textbf{0.8596}&\textbf{0.8615}&\textbf{0.0910}\\
			
			\textit{Liu2019PTPAMI} \cite{liu2019exploiting}&DU &0.8507&0.8451&0.1105 &0.8185&0.8100&0.0979 &0.8096&0.8269&0.0644 &0.8300&0.8121&0.1095 &0.7593&0.7346&0.1206 &0.8103&0.8140&0.1081\\
			
			\textit{Wang2022ACMMM} \cite{wang2022super}&MU &0.8657&0.8608&0.0870 &0.8210&0.8206&0.0854 &0.8265&0.8122&0.0747 &0.8764&0.8665&0.0862 &0.8311&0.8215&0.0994 &0.8383&0.8373&0.0973\\
			
			\textit{Madhusudana2022TIP} \cite{madhusudana2022image}&MU &0.9179&0.9254&0.0784 &0.7708&0.7656&0.0935 &0.7152&0.6569&0.1099 &0.8908&0.8842&0.1042 &0.8474&0.8579&0.1017 &0.8495&0.8531&0.0943\\
			
			\hline
			BMQA$_{_{\mathrm{RN50}\raisebox{-0.35mm}{+}\mathrm{TransF}}}^{image\raisebox{-0.35mm}{-}text}$&MU &0.9348&0.9187&0.0695 &0.9064&0.8813&0.0686 &0.9075&0.8712&0.0638 &0.9364&0.9140&0.0726 &0.9011&0.8860&0.0816 &0.8989&0.8877&0.0830 \\
			BMQA$_{_{\mathrm{ViT\raisebox{-0.35mm}{-}B32}\raisebox{-0.35mm}{+}\mathrm{TransF}}}^{image\raisebox{-0.35mm}{-}text}$&MU &0.9320&0.9264&0.0715 &\underline{\textbf{0.9131}}&\underline{\textbf{0.9042}}&0.0664 &\underline{\textbf{0.9108}}&\underline{\textbf{0.9067}}&\underline{\textbf{0.0598}} &\underline{\textbf{0.9488}}&0.9282&0.0737 &\underline{\textbf{0.9041}}&\underline{\textbf{0.8913}}&0.0828 &0.9089&0.9040&0.0816 \\
			BMQA$_{_{\mathrm{CNXT\raisebox{-0.35mm}{-}B}\raisebox{-0.35mm}{+}\mathrm{TransF}}}^{image\raisebox{-0.35mm}{-}text}$&MU &\underline{\textbf{0.9404}}&\underline{\textbf{0.9310}}&\underline{\textbf{0.0692}} &0.9128&0.9022&\underline{\textbf{0.0653}} &0.8920&0.8744&0.0607 &0.9464&\underline{\textbf{0.9358}}&\underline{\textbf{0.0717}} &0.8959&0.8850&\underline{\textbf{0.0808}} &\underline{\textbf{0.9121}}&\underline{\textbf{0.9065}}&\underline{\textbf{0.0802}}\\
			
			\bottomrule[1.5pt]        
\end{tabular}
\end{lrbox}
\scalebox{0.83}{\usebox{\tablebox}}	
}
\end{table*}

\subsection{Overall Performance Comparison}\label{subSec:overall_performance_comparison}
To demonstrate the overall quality forecasting performance, we compare our BMQA with 25 representative BIQA methods on the MLIQ database. For fairness, all comparison  methods use the same experimental settings as described in Sec.~\ref{subsec:experimental_protocols}. 
According to the type of quality feature extraction, these BIQAs are divided into two categories: 1) hand-crafted and 2) deep-learned methods.

\vspace{3pt}
\noindent \textbf{Hand-crafted BIQAs.}
Considering that low-light images on MLIQ are characterized by the hybrid multiple distortions, the comparison methods based on hand-crafted features are composed of four types:
\begin{itemize}
	\vspace{-3pt}
	\item \textbf{GP}:  \underline{g}eneral-\underline{p}ropose BIQAs, including \textit{Mittal2012TIP} \cite{mittal2012no}, \textit{Liu2014SPIC} \cite{liu2014no}, and \textit{Zhang2015TIP} \cite{zhang2015feature}.
	
	\item \textbf{MD}: distortion-specific BIQAs for \underline{m}ultiply \underline{d}istortion, including \textit{Li2016SPL} \cite{li2016no}.
	
	\item \textbf{CD}: distortion-specific BIQAs for \underline{c}ontrast \underline{d}istortion, including \textit{Gu2017TCB} \cite{gu2017no}.
	
	\item \textbf{LL}: distortion-specific BIQAs for \underline{l}ow-\underline{l}ight distortion, including \textit{Xiang2020TMM} \cite{xiang2020blind}, \textit{Wang2021ICME} \cite{wang2021blind},  and \textit{Wang2022TII} \cite{wang2022low}.
	\vspace{-3pt}
\end{itemize}
It is noted that the implementation of \textit{Xiang2020TMM} \cite{xiang2020blind} is not available, and we provide the results of our re-implementation. For a fair comparison, all the other  results in Table~\ref{tab:performance_comparison} are obtained by running the released implementations from the corresponding authors.

\vspace{3pt}
\noindent \textbf{Deep-learned BIQAs.}
Considering that low-light images on MLIQ are characterized by  some specific distortions, the comparison methods based on deep-learned features can be divided into four categories:
\begin{itemize}
	\vspace{-3pt}
	\item \textbf{AS}: \underline{a}llocation-based \underline{s}upervised BIQAs, including \textit{Kang2014CVPR} \cite{kang2014convolutional}, \textit{Bosse2018TIP} \cite{bosse2017deep}, and \textit{Ke2021ICCV} \cite{ke2021musiq}.
	
	\item \textbf{GS}: \underline{g}eneration-based \underline{s}upervised BIQAs, including \textit{Ying2020CVPR} \cite{ying2020patches}.

	\item \textbf{MT}: \underline{m}ulti-\underline{t}ask based supervised BIQAs, including \textit{Kim2019TNNLS} \cite{kim2019deep}, \textit{Yan2019TMM} \cite{yan2019naturalness}, \textit{Zhang2020TCSVT} \cite{zhang2020blind}.
	
	\item \textbf{MO}: \underline{m}ulti-\underline{o}bjective based supervised BIQAs,  including \textit{Wu2020TIP} \cite{wu2020end}, \textit{Li2020ACMMM} \cite{li2020norm}, \textit{Su2020CVPR} \cite{su2020blindly}, \textit{Zhu2020CVPR} \cite{zhu2020metaiqa}.
	
	\item \textbf{MC}: \underline{m}ulti-objective based supervised BIQAs with \underline{c}rossdataset-learned constraint, including \textit{Ma2021ACMMM} \cite{ma2021remember}, \textit{Zhang2021TIP} \cite{zhang2021uncertainty}, and \textit{Zhang2022TPAMI} \cite{zhang2022continual}.
	
	\item \textbf{MU}: \underline{m}etric-based \underline{u}nsupervised BIQAs, including \textit{Wang2022ACMMM} \cite{wang2022super} and \textit{Madhusudana2022TIP} \cite{madhusudana2022image}.
	
	\item \textbf{DU}: \underline{d}omain-based \underline{u}nsupervised BIQAs, including \textit{Liu2019TPAMI} \cite{liu2019exploiting}.
	\vspace{-3pt}
\end{itemize}

\noindent \textbf{Performance Comparison.}
Table \ref{tab:performance_comparison} tabulates the overall  results of $25$ representative BIQA methods. In addition, we provide the results of three BMQA variants, including
BMQA$_{_{\mathrm{RN\raisebox{-0.35mm}{-}50}\raisebox{-0.35mm}{+}\mathrm{TransF}}}^{image\raisebox{-0.35mm}{-}text}$, BMQA$_{_{\mathrm{ViT\raisebox{-0.35mm}{-}B32}\raisebox{-0.35mm}{+}\mathrm{TransF}}}^{image\raisebox{-0.35mm}{-}text}$ and BMQA$_{_{\mathrm{CNXT\raisebox{-0.35mm}{-}B}\raisebox{-0.35mm}{+}\mathrm{TransF}}}^{image\raisebox{-0.35mm}{-}text}$, which take TransF as $\mathcal{F}_{txt}$ and take RN-50, ViT-B32, and CNXT-B as $\mathcal{F}_{img}$, respectively.

From Table \ref{tab:performance_comparison}, it is observed that the average prediction accuracy of three BMQAs significantly outperforms the other $25$ competitive BIQA methods. 
The best BMQA$_{_{\mathrm{CNXT}\raisebox{-0.35mm}{+}\mathrm{TransF}}}^{image\raisebox{-0.35mm}{-}text}$ obtains a PLCC score of {$0.9121$, a SRCC score of $0.9065$, and a RMSE score of $0.0802$}, which is very promising in the BIQA task with authentic distortions. 
Furthermore, the experimental results also show that there is no significant performance difference for our BMQA from {Device-I to Device-V}, and the entire database, suggesting that different shooting devices have a slight side-effect in  multimodal learning.

\subsection{Further Discussion}
In this section, we mainly discuss the applicability and generalization of our BMQA$^{image\raisebox{-0.35mm}{-}only}$ method in the current mainstream of single-image modality scenario.

\subsubsection{Motivation}
From Sec. \ref{subSec:framework_generalization_validation} to \ref{subSec:overall_performance_comparison}, we have validated the effectiveness of our BMQA$^{image\raisebox{-0.35mm}{-}text}$ in terms of feature representation, ablation study, and overall performance. 
Nevertheless, the proposed BMQA has to face the challenge due to the absence of text description. In other words, the question is how to use our BMQA when the QSD-based text modality is absent. To address this challenge, we further propose a feasible scheme and validate it on another independent low-light database, Dark-4K, which contains only image samples and their MOS labels.

\subsubsection{Additional Modality Generation}
Considering that image quality can be mainly represented by the related language information in the BIQA task, we believe that $X_{txt}$ can also be replaced by a recent image caption model: `show and tell' (SAT) \cite{xu2015show}. Therefore, when the text modality is unavailable, BMQA employs an image captioning model to generate the feature representation of $X_{txt}$.

To the best of our survey, few existing image captioning models are specially trained for the BIQA task, and hence the generated caption is less relevant to the visual quality experience, which may not meet the text generation principles described in Sec. \ref{subSec:text_modality}. Therefore, we train a special QSD-based captioning model $\mathcal{F}_{cap}$ based on the MLIQ database (\textit{e.g.}, 8:1:1 partition). 
\begin{figure}[!t]
\centering
\begin{tabular}{p{2.5cm}<{\centering} p{2.5cm}<{\centering} p{2.5cm}<{\centering} }
\includegraphics[width=1.13\linewidth, height=.6\linewidth]{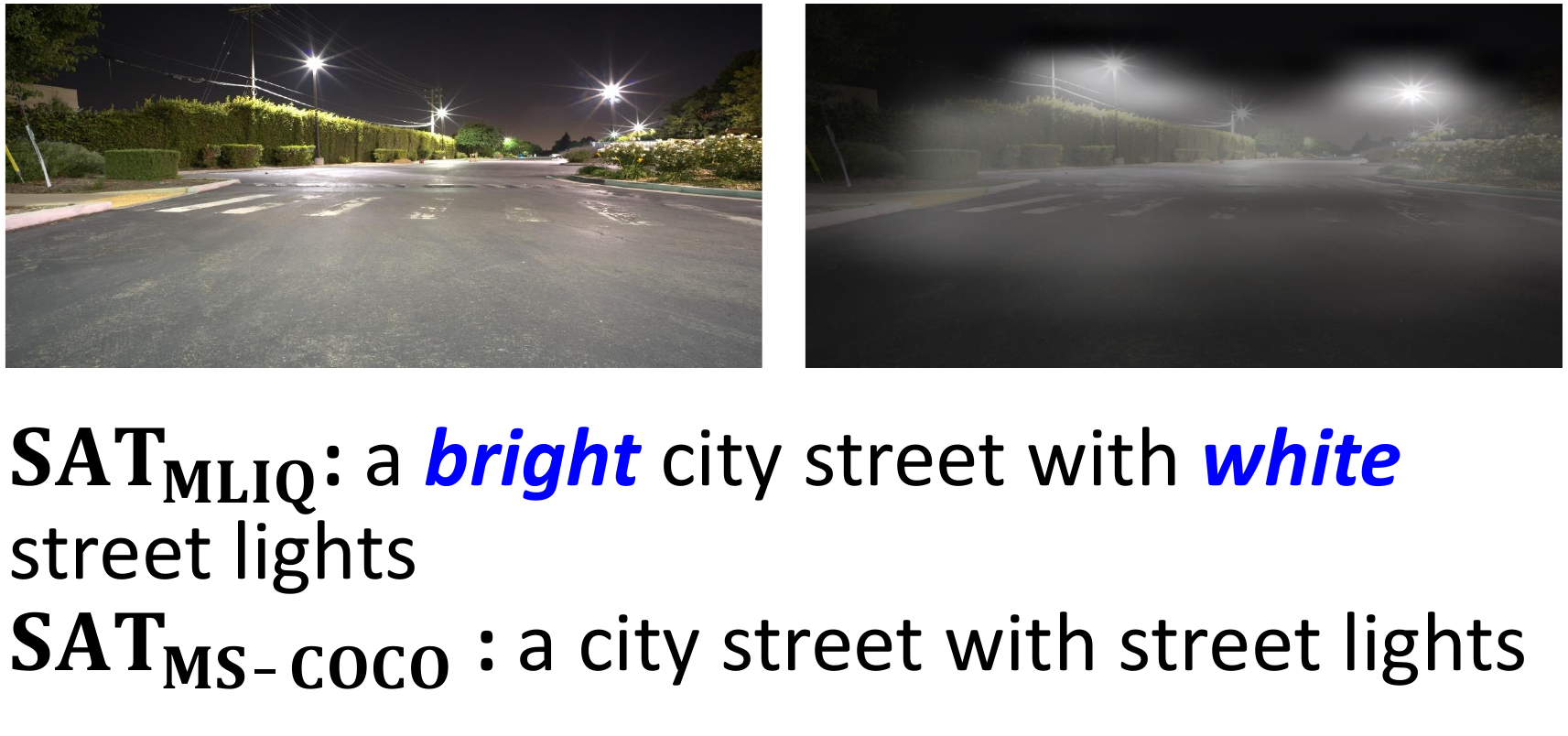} 
&\includegraphics[width=1.13\linewidth, height=.6\linewidth]{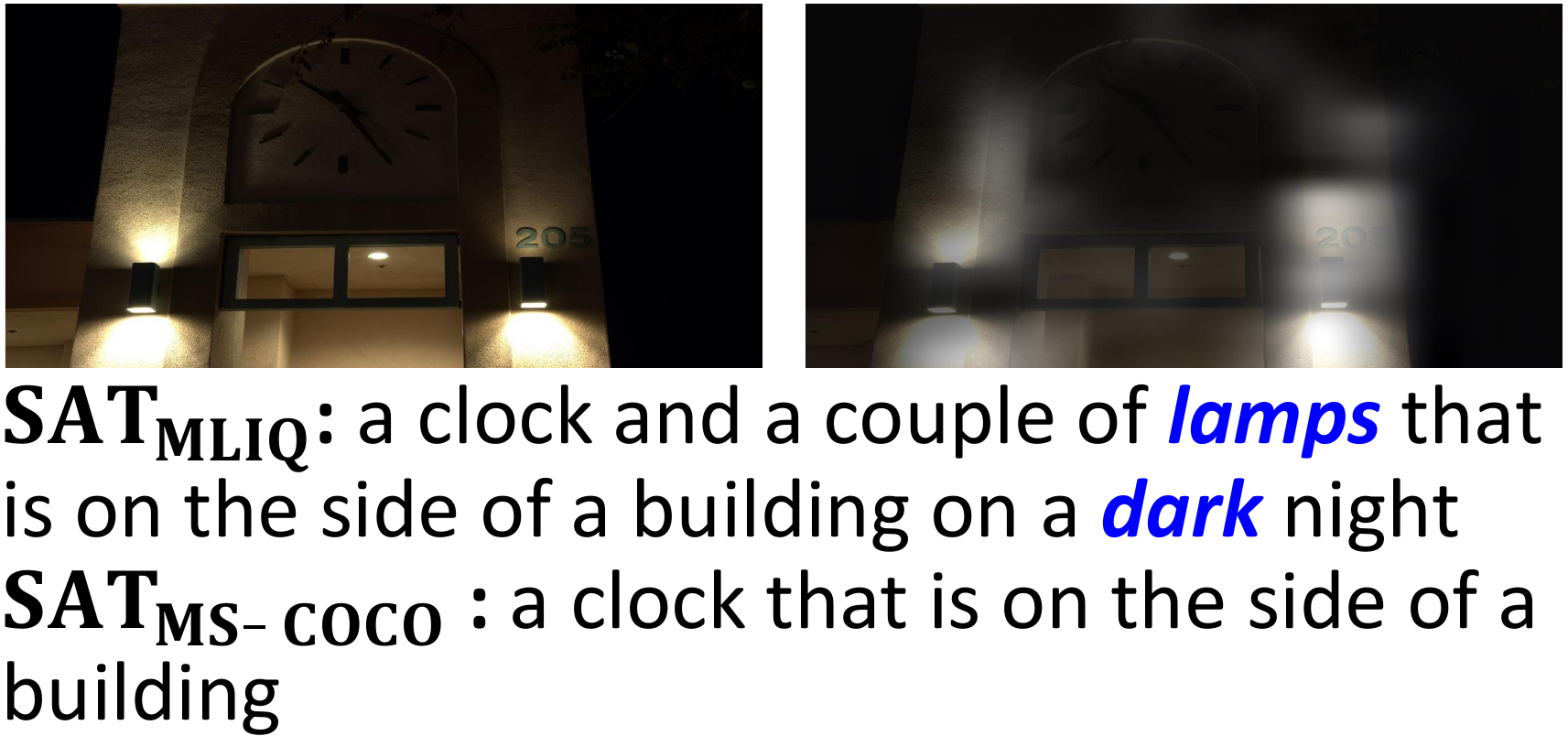} 
&\includegraphics[width=1.13\linewidth, height=.6\linewidth]{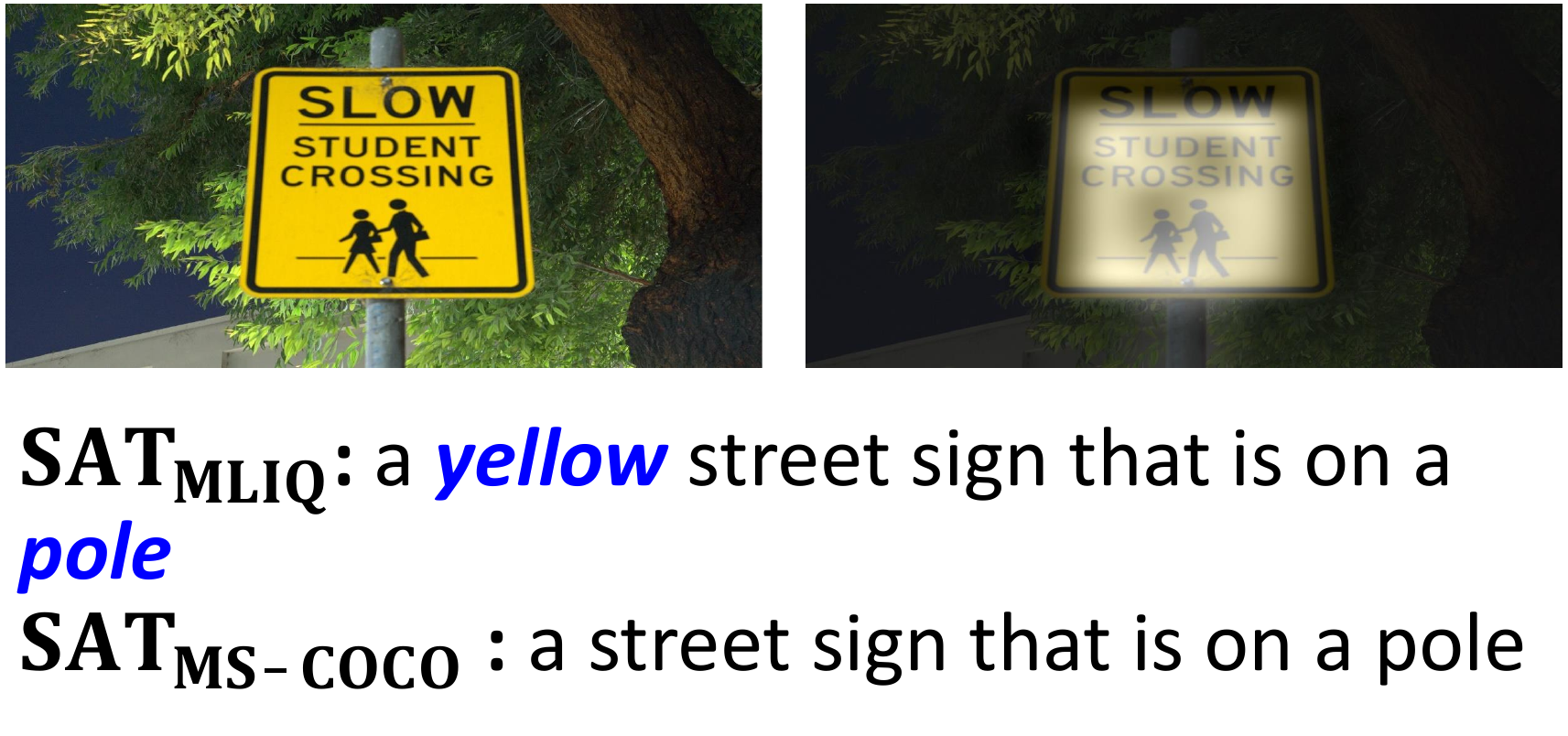}  \\   	
\includegraphics[width=1.13\linewidth, height=.6\linewidth]{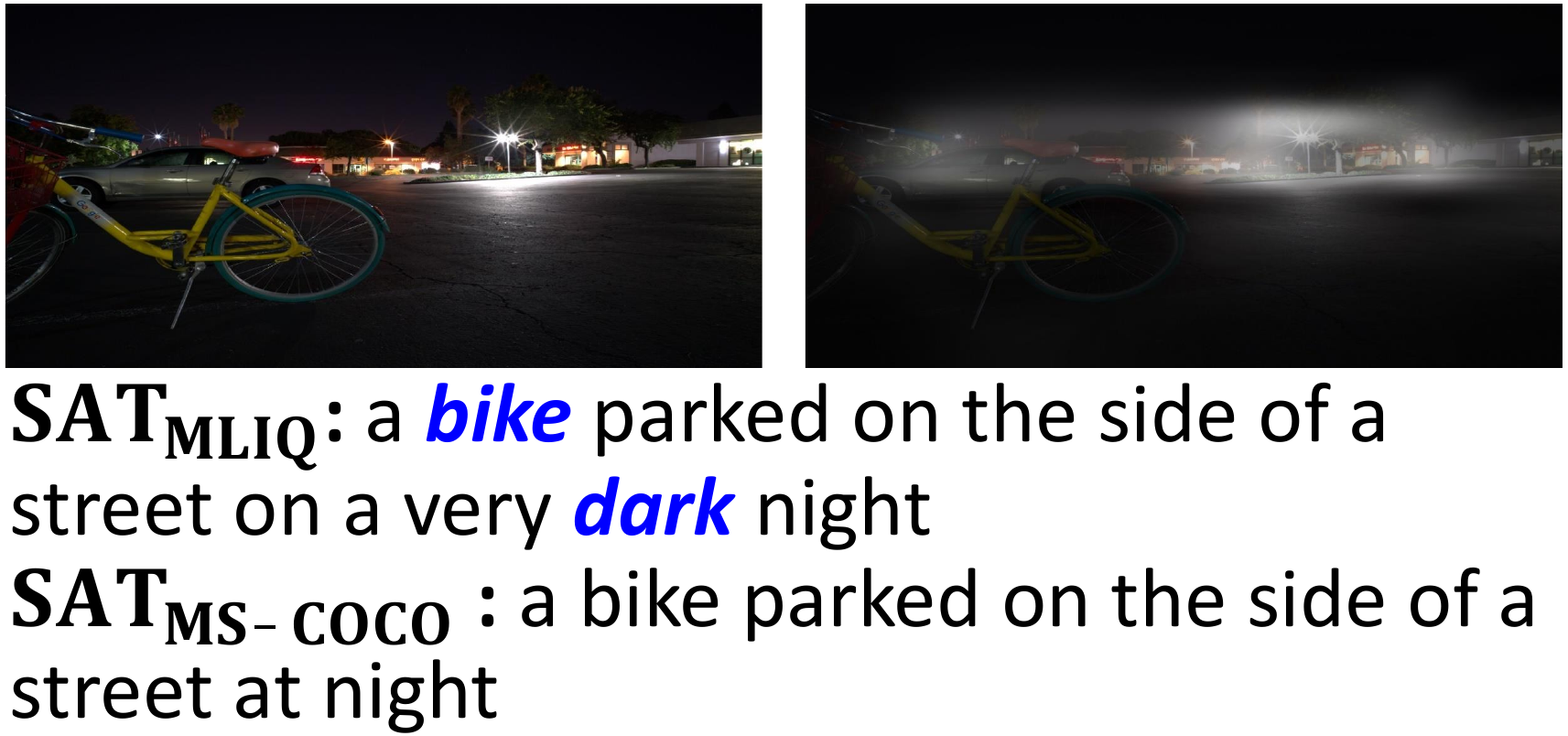} 
&\includegraphics[width=1.13\linewidth, height=.6\linewidth]{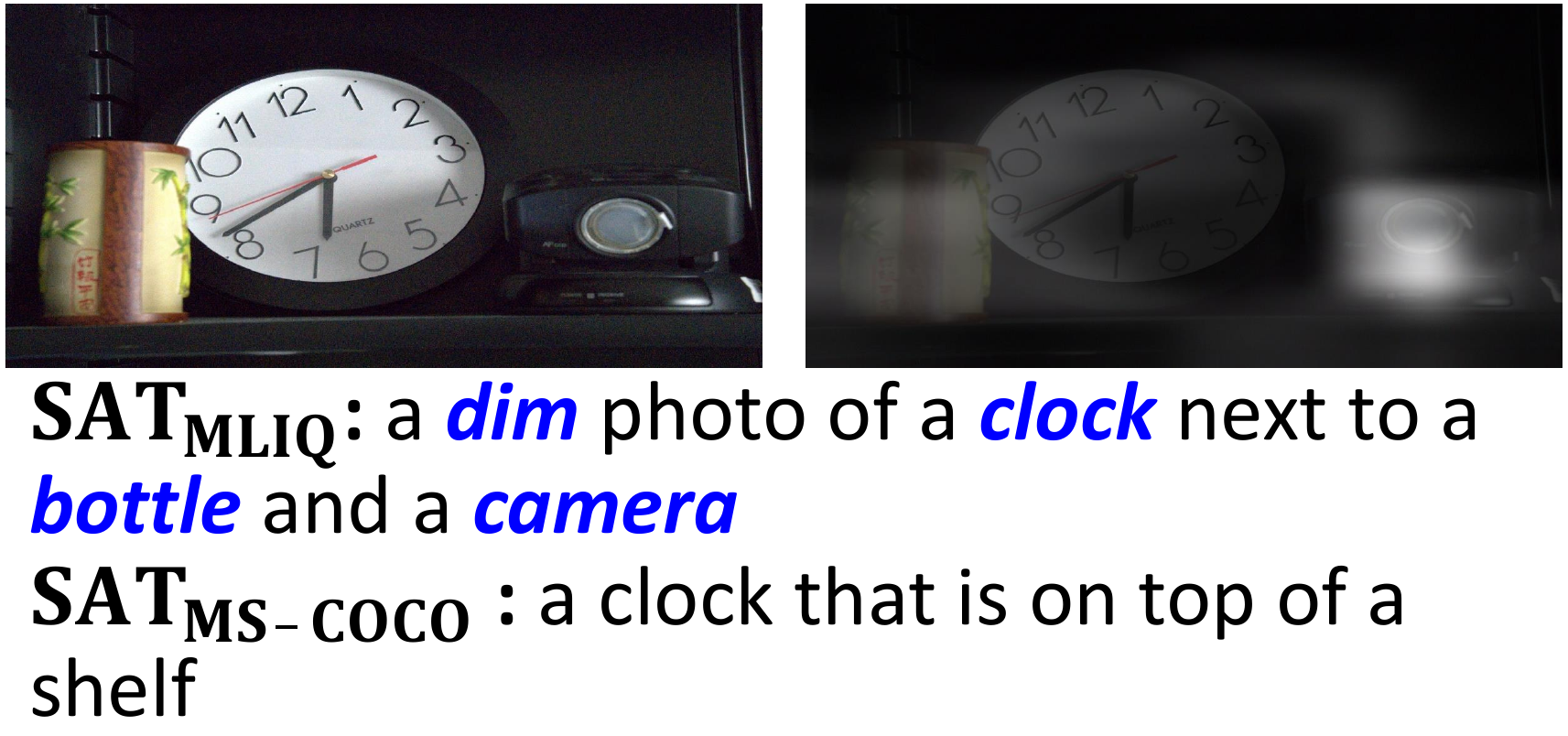} 
&\includegraphics[width=1.13\linewidth, height=.6\linewidth]{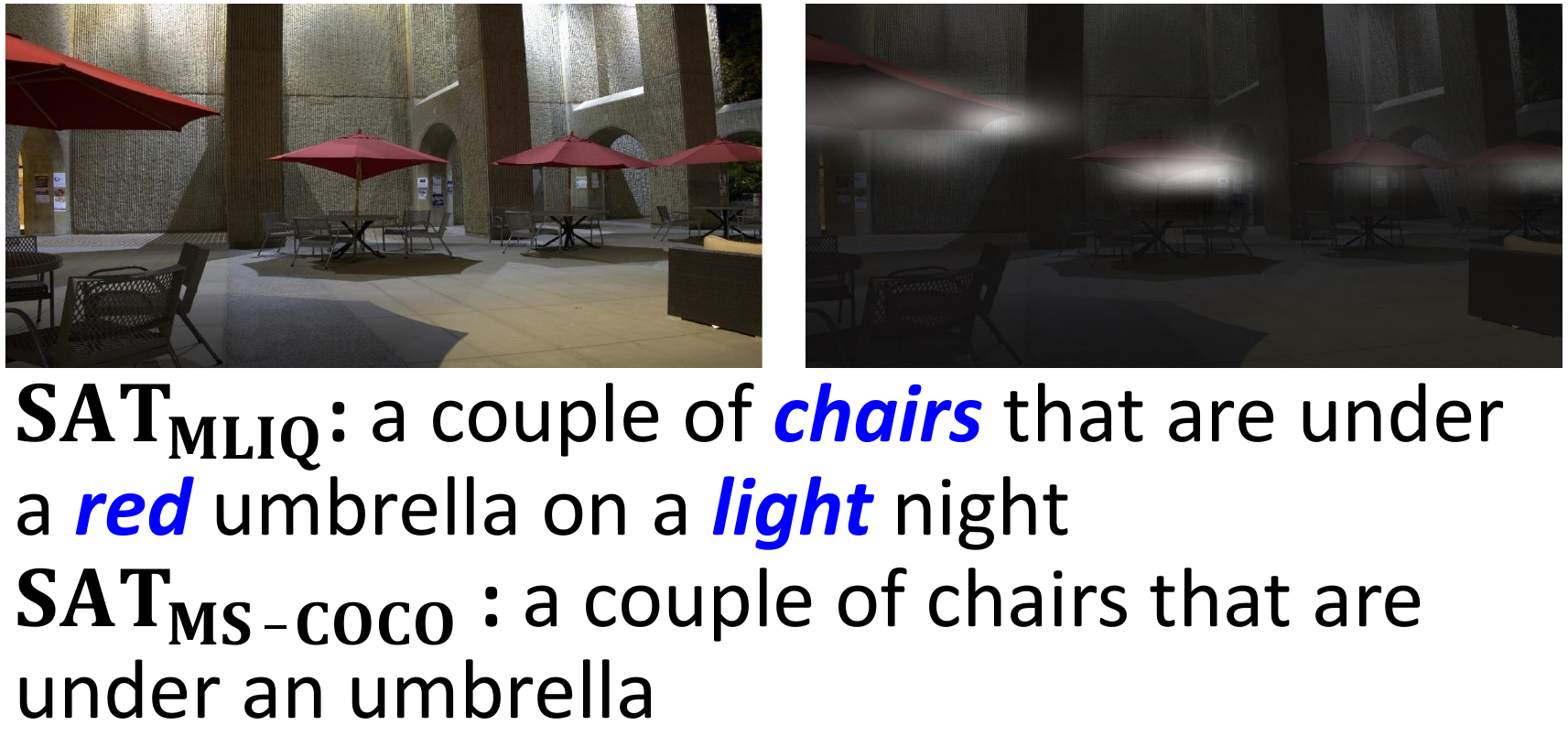}  \\   
\end{tabular}   
	\caption{\textbf{Examples of the generated text modality on our Dark-4K database}. Each example contains an original image sample (\textit{left}), an attention map of the quality cognition (\textit{right}), a predicted caption result  obtained by a pre-trained SAT caption model ($\mathrm{\mathbf{SAT}}_{\mathrm{\mathbf{MS\raisebox{-0.30mm}{-}COCO}}}$) on the MS-COCO database, and a low-light QSD-based caption result of $\mathrm{\mathbf{SAT}}_{\mathrm{\mathbf{MLIQ}}}$ trained on the MLIQ database.}
\label{fig:visualization_text_generation_examples}
\end{figure}

Specifically, each image sample is fed into $\mathcal{F}_{cap}$, and the associated text description is used as the training label. The learning goal is to maximize the joint pair-wise probability of the predicted caption and the annotated caption, or equivalently to minimize the log-likelihood over the training set as:
\begin{equation}\label{eqn:image_caption_loss}
\small
\vspace{-3pt}
\resizebox{.87\hsize}{!}{$	
\boldsymbol{\min}\mathbb{E}\left\{-\log\sum\limits_{1<k<N} \mathcal{P}\left[\mathcal{F}_{cap}\left(X_{img}{;~}\boldsymbol{\theta}_{cap}\right), {x_{k}}\right]\right\}
$},
\end{equation}
where $\boldsymbol{\theta}_{cap}$ represents the model weights of $\mathcal{F}_{cap}$.

Therefore, given an image $X_{img}$, the text modality is generated by 
\begin{equation}\label{eqn:text_generation}
\small
\vspace{-3pt}
<x^{_{'}}_{1},\cdots,x^{_{'}}_{N}>=\mathcal{F}_{cap}\left(X_{img}{;~}\boldsymbol{\theta}_{cap}\right),
\end{equation}
where {\small$<x^{_{'}}_{1},\cdots,x^{_{'}}_{N}>$} denotes the predicted QSD-based text modality.

To verify the feasibility of $\mathcal{F}_{cap}$,  we directly adopt the SAT model as the backbone of $\mathcal{F}_{cap}$. In the experiments, we employ a pre-trained model provided by \cite{xu2015show} (\textit{denoted as} SAT$_{MS-COCO}$), and finetune a special QSD-based captioning model on the MLIQ database for 100 epochs (\textit{denoted as} SAT$_{MLIQ}$). Finally, the TOP-5 accuracy of caption matching performance on the testing set reaches {$89.74\%$, and the {BLEU-4} score reaches $29.36\%$}.

Fig. \ref{fig:visualization_text_generation_examples} shows some QSD-based captioning results of SAT$_{MLIQ}$ and SAT$_{MS-COCO}$ on the Dark-4K database. As seen, SAT$_{MLIQ}$ contains more quality words related to the visual experience, such as brightness status, salient objects, and color status.  
To further explore the relationship between these keywords and the original images, we provide the attention maps corresponding to these keywords. As seen, the region of interest (ROI) is consistent with the quality keywords. Therefore, we believe that a well-trained QSD-based captioning model is competent to generate the text modality provided in MLIQ.

\subsubsection{Cross-dataset Validation on Image-only Case}
To verify the effect of $\mathcal{F}_{cap}$ in Eq. \eqref{eqn:text_generation}, we conduct the cross-database validation on the Dark-4K database. 
For a fair comparison, all BIQAs including our BMQA models have not been {fine-tuned or retrained}. In other words, all trained models are kept exactly the same as in Sec.~\ref{subSec:overall_performance_comparison}. It is worth noting  that since $\mathcal{F}_{cap}$ does not exist on the Dark-4K database, a specially-learned model $\mathcal{F}_{cap}$ is used to replace $\mathcal{F}_{cap}$ in \textbf{Algorithm} \ref{algm:bmqa}. Apart from this, other settings are the same as in Sec. \ref{subSec:training_detail}.

Table \ref{tab:performance_comparison_dark4k} provides the overall comparison results of all {$17$ deep-learned} BIQAs in terms of PLCC, SRCC, and RMSE. As seen, our BMQAs achieve the state-of-the-art performance on the cross-dataset validation. 
The best BMQA$_{_{\mathrm{CNXT\raisebox{-0.35mm}{-}B}\raisebox{-0.35mm}{+}\mathrm{TransF}}}^{image\raisebox{-0.35mm}{-}only}$ obtains a PLCC score of {$0.9156$, a SRCC score of $0.9085$, and a RMSE score of $0.0605$}. Experimental results verify the applicability and generalization performance of our BMQA framework.
\begin{table}[!t]
\vspace{-0.4cm}
\caption{\textbf{Performance comparison of the proposed BMQA and $17$ deep-learned BIQAs on the Dark-4K database}. The best results of the deep-learned BIQAs are highlighted in \textbf{bold}, and the best results of our BMQA are highlighted in \underline{\textbf{underline}}.}
\label{tab:performance_comparison_dark4k}
\centering
\scriptsize
\renewcommand{\arraystretch}{1.15}
\setlength{\tabcolsep}{2.8mm}{	
\begin{tabular}{rc|rrr}
			\toprule[1.5pt]
			\multicolumn{2}{c|}{Method Type} & \multicolumn{3}{c}{Entire Database}\\
			
			\Xhline{1.0pt}
			\rowcolor[gray]{.95} Deep-learned BIQAs &Type &PLCC $\uparrow$&SRCC $\uparrow$&RMSE $\downarrow$ \\

			\textit{Kang2014CVPR} \cite{kang2014convolutional} &AS &0.5434&0.5605&0.1307\\
			\textit{Bosse2018TIP} \cite{bosse2017deep} &AS &0.6552&0.6442&0.1028\\ 
			\textit{Ke2021ICCV} \cite{ke2021musiq} &AS &\textbf{0.8630}&\textbf{0.8382}&0.1195\\
			
			\textit{Ying2020CVPR} \cite{ying2020patches} &GS &0.6453&0.5802&0.0631\\
			
			\textit{Kim2019TNNLS} \cite{kim2019deep} &MT &0.5968&0.6030&0.1205\\
			\textit{Yan2019TMM} \cite{yan2019naturalness} &MT	 &0.5588&0.5610&0.1322\\
			\textit{Zhang2020TCSVT} \cite{zhang2020blind} &MT &0.5187&0.5581&0.1346\\
			
			\textit{Wu2020TIP} \cite{wu2020end} &MO &0.6216&0.5877&0.1248\\
			\textit{Li2020ACMMM} \cite{li2020norm} &MO &0.6138&0.6358&0.1107\\
			\textit{Su2020CVPR} \cite{su2020blindly} &MO
			&0.7914&0.7829&0.0755\\
			\textit{Zhu2020CVPR} \cite{zhu2020metaiqa} &MO &0.7791&0.7616&0.0931\\
			
			\textit{Ma2021ACMMM} \cite{ma2021remember} &MC &0.7713&0.7419&0.0810\\
			\textit{Zhang2021TIP} \cite{zhang2021uncertainty} &MC &0.8079&0.8131&0.0697\\
			\textit{Zhang2022TPAMI} \cite{zhang2022continual} &MC &0.8538&0.8358&\textbf{0.0593}\\
			
			\textit{Liu2019TPAMI} \cite{liu2019exploiting} &DU &0.6319&0.6352&0.1002\\
			\textit{Wang2022ACMMM} \cite{wang2022super} &MU &0.8526&0.8357&0.0687\\
			\textit{Madhusudana2022TIP} \cite{madhusudana2022image} &MU &0.8293&0.8240&0.0669\\

			\hline
			BMQA$_{_{\mathrm{RN}\raisebox{-0.35mm}{+}\mathrm{TransF}}}^{image\raisebox{-0.35mm}{-}only}$&MU &0.9026&0.8874&0.0624\\
			BMQA$_{_{\mathrm{ViT\raisebox{-0.35mm}{-}B32}\raisebox{-0.35mm}{+}\mathrm{TransF}}}^{image\raisebox{-0.35mm}{-}only}$&MU &0.9117&0.8989&\underline{\textbf{0.0593}}\\
			BMQA$_{_{\mathrm{CNXT\raisebox{-0.35mm}{-}B}\raisebox{-0.35mm}{+}\mathrm{TransF}}}^{image\raisebox{-0.35mm}{-}only}$&MU &\underline{\textbf{0.9156}}&\underline{\textbf{0.9085}}&0.0605\\
			
\bottomrule[1.5pt]        
\end{tabular}}
\end{table}

\section{Conclusion and Future Work}
In this article, we have presented a new study on blind multimodal quality assessment (BMQA) of low-light images from both subjective and objective perspectives. Specifically, we establish the first multimodal quality assessment database for low-light images, where two quality-aware principles are designed for multimodal benchmark construction.
Moreover, we investigate the BMQA framework by exploring quality feature representation, alignment and fusion, and multimodal learning. 
Experimental results verify the effectiveness of our BMQA  compared with state-of-the-art methods on two different low-light benchmark databases. We believe that this work can bring a new research perspective to image and video quality assessment.

In the future, BMQA can be further studied and verified in many aspects. First, we demonstrate how multimodal data is able to improve the performance of blind image quality assessment (BIQA) on the low-light application scenario, rather than how multimodal data collectively determines the visual quality experience. The latter is more challenging and thus requires deploying more sensors and fusing multi-sensor data.

Second,  we design two quality semantic description (QSD) principles and only consider some QSD-based text features in multimodal learning. However, image quality perception can be very complex and hence more interpreting attributes (\textit{e.g.}, sharpness, composition, and aesthetic.) need to be further investigated.

Third, we employ the multimodal self-supervision and supervision mechanisms to train our baseline model. The current design is simple and efficient on low-light images, but the interaction between image and text modalities needs to be further explored for the BMQA framework, which will help to extract more efficient feature representations.

Last but not least, we construct a multimodal low-light image database and present an effective no-reference quality indicator because it is more challenging and urgent. However, the feasibility of multimodality-driven quality assessment needs to be verified on more benchmark databases, including those for distortion-specific and general-purpose applications. Furthermore, its feasibility needs to be studied and verified on the full-reference task, as this quality assessment paradigm is still used in many real-world scenarios.


{
\balance
\bibliographystyle{ieee_fullname}
\bibliographystyle{IEEEtran}
\bibliography{mmqa2022bib}
}

\end{document}